\documentclass[sigconf]{acmart}
\usepackage{graphicx}
\usepackage{amsmath}

\usepackage{amssymb}

\usepackage{booktabs}
\usepackage{color}
\definecolor{mygreen}{RGB}{0,176,80} 
\definecolor{myred}{RGB}{192,0,0} 
\usepackage{multirow}
\usepackage{bm}
\usepackage{enumitem}
\usepackage{makecell}
\usepackage{tabularx}
\usepackage{balance}
\usepackage{amssymb}

\AtBeginDocument{%
  \providecommand\BibTeX{{%
    \normalfont B\kern-0.5em{\scshape i\kern-0.25em b}\kern-0.8em\TeX}}}

\copyrightyear{2023}
\acmYear{2023}
\setcopyright{acmlicensed}\acmConference[MM '23]{Proceedings of the 31st ACM International Conference on Multimedia}{October 29-November 3, 2023}{Ottawa, ON, Canada}
\acmBooktitle{Proceedings of the 31st ACM International Conference on Multimedia (MM '23), October 29-November 3, 2023, Ottawa, ON, Canada}
\acmPrice{15.00}
\acmDOI{10.1145/3581783.3611768}
\acmISBN{979-8-4007-0108-5/23/10}

\newcommand{\short}{\textsc{Beat}}
\newcommand{\module}{\textsc{REM-G}}
\newcommand{\longname}{\textbf{B}i-directional one-to-many \textbf{E}mbedding \textbf{A}lignmen\textbf{t}}

\begin{document}

\title{\short{}: Bi-directional One-to-Many Embedding Alignment for  Text-based Person Retrieval }

\author{Yiwei Ma}
\email{yiweima@stu.xmu.edu.cn}


\affiliation{
  \institution{Key Laboratory of Multimedia Trusted Perception and Efficient Computing,\\
  Ministry of Education of China,\\
  Xiamen University,}
  \city{Xiamen}
  \state{Fujian}
  \country{China}
}

\author{Xiaoshuai Sun}
\email{xssun@xmu.edu.cn}
\authornote{Corresponding author.}
\affiliation{
  \institution{Key Laboratory of Multimedia Trusted Perception and Efficient Computing,\\
  Ministry of Education of China,\\
  Xiamen University,}
  \city{Xiamen}
  \state{Fujian}
  \country{China}
}

\author{Jiayi Ji}
\email{ jjyxmu@gmail.com}
\affiliation{
  \institution{Key Laboratory of Multimedia Trusted Perception and Efficient Computing,\\
  Ministry of Education of China,\\
  Xiamen University,}
  \city{Xiamen}
  \state{Fujian}
  \country{China}
}

\author{Guannan Jiang}
\email{jianggn@catl.com}
\affiliation{
  \institution{Contemporary Amperex Technology Co., Limited}
  \city{Ningde}
  \state{Fujian}
  \country{China}
}

\author{Weilin Zhuang}
\email{ ZhuangWL@catl.com}
\affiliation{
  \institution{Contemporary Amperex Technology Co., Limited}
  \city{Ningde}
  \state{Fujian}
  \country{China}
}

\author{Rongrong Ji}
\email{rrji@xmu.edu.cn}
\affiliation{
  \institution{Key Laboratory of Multimedia Trusted Perception and Efficient Computing,\\
  Ministry of Education of China,\\
  Xiamen University,}
  \city{Xiamen}
  \state{Fujian}
  \country{China}
}

\renewcommand{\shortauthors}{Yiwei Ma et al.}

\begin{abstract}

Text-based person retrieval (TPR) is a challenging task that involves retrieving a specific individual based on a textual description. Despite considerable efforts to bridge the gap between vision and language, the significant differences between these modalities continue to pose a challenge. Previous methods have attempted to align text and image samples in a modal-shared space, but they face uncertainties in optimization directions due to the movable features of both modalities and the failure to account for one-to-many relationships of image-text pairs in TPR datasets.
To address this issue, we propose an effective bi-directional one-to-many embedding paradigm that offers a clear optimization direction for each sample, thus mitigating the optimization problem. Additionally, this embedding scheme generates multiple features for each sample without introducing trainable parameters, making it easier to align with several positive samples. Based on this paradigm,  we propose a novel \longname{} (\short{}) model to address the TPR task.
Our experimental results demonstrate that the proposed \short{} model achieves state-of-the-art performance on three popular TPR datasets, including CUHK-PEDES (\emph{65.61 R@1}), ICFG-PEDES (\emph{58.25 R@1}), and RSTPReID (\emph{48.10 R@1}). Furthermore, additional experiments on MS-COCO, CUB, and Flowers datasets further demonstrate the potential of \short{} to be applied to other image-text retrieval tasks.

\end{abstract}



\maketitle

\begin{figure}
\centering 
  \includegraphics[width=1.0\columnwidth]{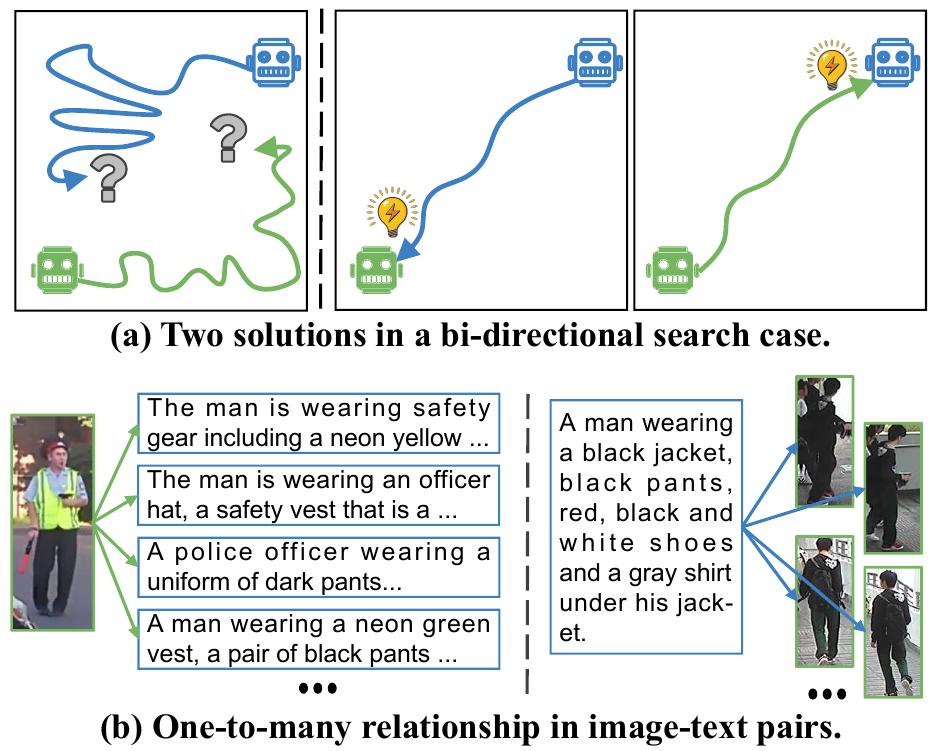}

  \vspace{-0.3cm}
  \caption{ (a) Two possible approaches for the bi-directional search problem. The left-hand side illustrates both agents moving toward each other to meet at a midpoint, while the right-hand side shows one agent remaining stationary while the other is movable. (b) A one-to-many relationship of image-text pairs in the TPR dataset.
  }

  \vspace{-0.3cm}
  \label{fig:intro}
\end{figure}

\section{Introduction}
\label{sec:intro}

The task of text-based person retrieval (TPR)~\cite{li2017person} is of critical importance as it aims to seek a specific person from a vast image gallery through a natural language description query. Unlike traditional image-based or attribute-based person retrieval methods, TPR leverages language descriptions that are more easily accessible and provide more comprehensive information. However, due to the significant difference between the visual and linguistic modalities, TPR is also a more challenging task than image-based or attribute-based person retrieval.

To address the significant modality gap, numerous previous works~\cite{li2017person,zhu2021dssl,han2021text} have adopted a modal-shared one-to-one embedding paradigm. This paradigm involves embedding images and textual descriptions into a joint space through a one-to-one projection function, which is then used to investigate semantic alignment between image and text pairs. While this paradigm is theoretically elegant and has been widely used for decades, it suffers from two limitations in optimization and alignment.

Firstly, the modal-shared embedding technique in this paradigm presents a significant challenge in optimizing the model. This is due to the fact that both visual and textual features are trainable, leading to fluctuations in the optimization direction. This can be likened to a bi-directional search scenario. As illustrated in Fig.~\ref{fig:intro}(a), when both agents are movable, it is difficult for them to find each other due to the uncertain meeting location. Similarly, when features of both modalities are movable during alignment, the uncertainty in optimization direction becomes even more pronounced. Consequently, the conventional modal-shared embedding paradigm is less effective than alternative solutions that fix one modality and only optimize the other.

\begin{figure*}
\centering 
  \includegraphics[width=2.1\columnwidth]{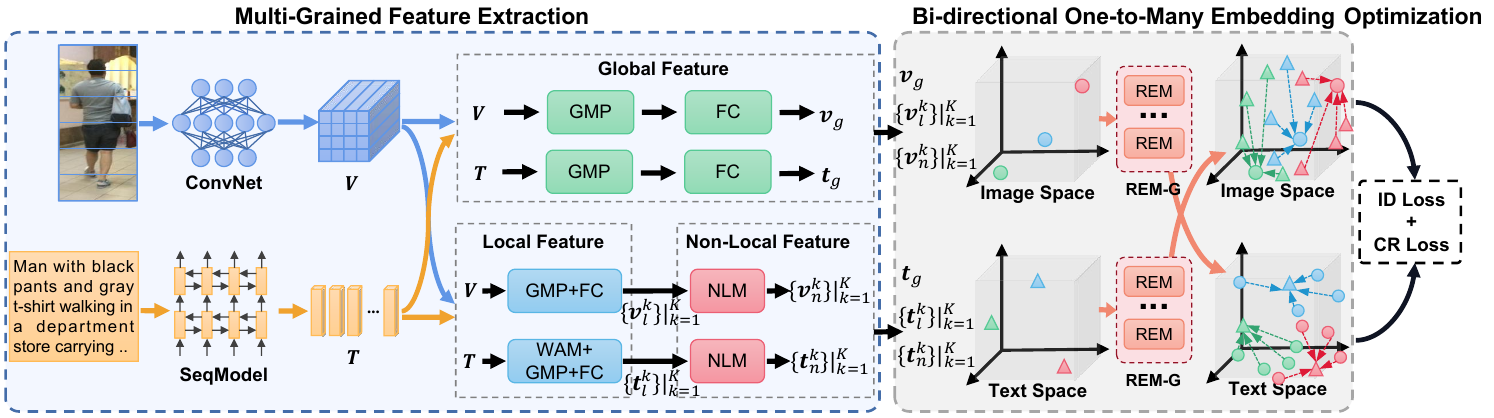}
  \vspace{-0.6cm}
  \caption{  Illustration of the proposed \short{} model. Images and texts are first processed by ConvNet and SeqModel to obtain visual features $\bm{V}$ and textual features $\bm{T}$, respectively. Global features $\boldsymbol{v}_g$ and $\boldsymbol{t}_g$ are extracted through a global max pooling (GMP) and a fully connected layer (FC) based on $\bm{V}$ and $\bm{T}$. Similarly, we adopt a global max pooling (GMP) and an FC layer to obtain $K$ local visual features $\{\boldsymbol{v}_l^k\}|_{k=1}^K$. Then, we employ word attention module (WAM)~\cite{ding2021semantically}, GMP and FC layer to obtain $K$ local textual features $\{\boldsymbol{t}_l^k\}|_{k=1}^K$. We then adopt a non-local module (NLM) to get $K$ non-local visual features $\{\boldsymbol{v}_n^k\}|_{k=1}^K$ and textual features $\{\boldsymbol{t}_n^k\}|_{k=1}^K$. Afterward, the \module{} is introduced to perform bi-directional one-to-many embedding. Finally, textual and visual samples are aligned in two modal-specific spaces under the guidance of ID loss and CR loss.
  }
  \vspace{-0.3cm}
  \label{fig:overview}
\end{figure*}

To address the optimization challenge of the modal-shared embedding paradigm, we propose a novel and effective approach called the \emph{bi-directional embedding paradigm}. This approach involves separately projecting images and text into modality-specific spaces for matching, as illustrated in Fig.~\ref{fig:overview}.
We first extract multi-grained textual and visual features using established methods~\cite{ding2021semantically}. Unlike the conventional modal-shared embedding paradigm that aligns both visual and textual features in a modal-shared space where both modalities are trainable, the proposed bi-directional embedding paradigm aligns samples in both image and text spaces. Specifically, we use a projection function to embed visual features into the text space and align them with textual features based on cosine similarity, where textual features are fixed in the text space. Similarly, the same operation is performed for textual features. This approach enables us to achieve effective alignment between the two modalities while mitigating the optimization problem that arises when both modalities are trainable.

Secondly, the one-to-one embedding scheme employed in prior studies~\cite{li2017person,zhu2021dssl,han2021text,FeiMatchStruICML22} inadequately accounts for the one-to-many relationship inherent in image-text pairs. This issue is demonstrated in Fig.~\ref{fig:intro}(b), where an image of a person may be associated with a varied range of textual descriptions that emphasize different aspects of the individual or depict them in diverse styles. Likewise, a textual description may correspond to multiple images of the same person due to the possibility of being captured by different cameras. This phenomenon presents a significant challenge in aligning a query with multiple positive samples since the traditional one-to-one embedding paradigm only maps the query to a single feature.

To tackle the challenge of aligning one query sample with multiple positive samples, we propose a novel approach called the \emph{one-to-many embedding paradigm}, which provides multiple embedding features for each sample. Our approach utilizes a residual embedding module group (\module{}), which comprises several residual embedding modules (REMs). Each REM performs one-to-one projection, enabling the \module{} to perform one-to-many embedding. It is worth noting that due to the bottleneck structure used for each REM, the one-to-many embedding does not introduce additional parameters. By generating multiple embedding features for each query sample, our approach allows several positive samples to align with the most suitable feature, thereby alleviating the difficulty of one-to-many alignment.


Based on the proposed bi-directional one-to-many embedding paradigm, we introduce a new model called the \longname{} (\textbf{\short{}}) model to tackle the challenging TPR task. To evaluate the proposed approach, we conduct extensive experiments on three TPR datasets: CUHK-PEDES~\cite{li2017person}, ICFG-PEDES~\cite{ding2021semantically}, and RSTPReID~\cite{zhu2021dssl}. Our results consistently demonstrate that the proposed \short{} model outperforms existing state-of-the-art (SOTA) approaches significantly. Moreover, our proposed paradigm also yields benefits on other image-text retrieval datasets, such as MS-COCO~\cite{lin2014microsoft}, CUB~\cite{reed2016learning}, and Flowers~\cite{reed2016learning}.

In summary, the main contributions of our work include:

\begin{itemize}[leftmargin=*, itemsep=2pt,topsep=0pt,parsep=0pt]
    \item We present a novel approach that uses a bi-directional embedding paradigm to align visual and textual samples in TPR, which addresses the issue of uncertain optimization direction in the traditional paradigm.
    
    \item We notice that the images and their corresponding representations in the TPR dataset exhibit a one-to-many relationship. Hence, we propose a  one-to-many embedding paradigm that models this relationship in image-text pairs, thereby improving one-to-many alignment for TPR.

    \item Building on the proposed bi-directional one-to-many embedding paradigm, we propose the \short{} model, which achieves SOTA performance on three TPR datasets and consistently improves other image-text retrieval tasks.

\end{itemize}

\section{Related Work}

\subsection{Text-based Person Retrieval}

The text-based person retrieval (TPR) task~\cite{xu2023mining,jiang2023cross,jiang_2023,wang2023exploiting,chen2023towards,wang2022caibc,wang2022sum,wang2022aspd}, which retrieves a person through natural language descriptions, differs from conventional image-based or attribute-based methods. Considerable research has been dedicated to this task, with Li \emph{et al.}~\cite{li2017person} proposing a CNN-LSTM network with gated neural attention, and Li \emph{et al.}~\cite{li2017identity} introducing an identity-aware two-stage framework to improve retrieval performance. Other approaches include Jing \emph{et al.}~\cite{jing2020pose}, who leverage pose information as a soft attention signal to locate important and discriminative regions, and Niu \emph{et al.}~\cite{niu2020improving}, who propose a multi-granularity image-text alignments model to mitigate the cross-modal fine-grained problem. Liu \emph{et al.}~\cite{liu2019deep} introduce a novel A-GANet to exploit semantic scene graphs, while Zhu \emph{et al.}~\cite{zhu2021dssl} propose deep surroundings-person separation learning to extract and match person information while reducing interference from surrounding information. Jing \emph{et al.}~\cite{jing2020cross} propose a moment alignment network to solve cross-modal cross-domain person search, while Wu \emph{et al.}~\cite{wu2021lapscore} adopt two color reasoning sub-tasks to improve model sensitivity to fine-grained cross-modal information.

All previous TPR methods employ the modal-shared one-to-one embedding paradigm, where image and text are embedded into a unified space for similarity evaluation via a one-to-one projection function. In this paper, we propose a novel bi-directional one-to-many embedding paradigm that reduces optimization difficulty and adapts to the one-to-many relationship in image-text pairs.

\subsection{Visual-Textual Embedding}
The field of visual-textual embedding technology is gaining increasing attention from researchers due to the rapid growth of vision and language data. In particular, several methods have been proposed to align visual and textual samples. Frome \emph{et al.}~\cite{frome2013devise} proposed a visual semantic embedding (VSE) framework that aligns images with sentences using ranking loss. Faghri \emph{et al.}~\cite{faghri2017vse++} improved the VSE model by focusing on computing the hardest negative examples, with the aim of improving computational efficiency. Huang \emph{et al.}~\cite{huang2017instance} introduced a novel multi-modal context-modulated attention scheme, implemented using a multi-modal LSTM network, to attend to an image-text pair at each timestep. Lee \emph{et al.}~\cite{lee2018stacked} proposed the stacked cross attention network (SCAN) which employs a fine-grained attention mechanism to embed images and sentences.
Aside from the conventional image-text matching task, visual-textual embedding has also benefited many other multi-modal tasks, such as image captioning (IC)\cite{ji2022konowing,ma2022knowing,Hu_2022_CVPR,fang2022injecting} and visual question answering (VQA)\cite{Jing_2022_CVPR,ding2022mukea,cascante2022simvqa,chun2021probabilistic}. For example, Anderson \emph{et al.}~\cite{anderson2018bottom} proposed bottom-up top-down attention for IC and VQA, while Jiang \emph{et al.}~\cite{jiang2020defense} explored the grid representation of the object detector to further improve the performance of VQA.

In this context, our work proposes a bi-directional one-to-many embedding paradigm for visual-textual embedding. This approach is complementary to the aforementioned works and has the potential to enhance the performance of a range of multi-modal tasks~\cite{wang2022look,fei-etal-2023-scene,lin2020many,wang2023towards}.

\section{The Proposed Method}

In this section, we elaborate on each component of the proposed \short{}, which is illustrated in Fig.~\ref{fig:overview}. Specifically, we first introduce the problem formulation of the TPR task in Sec.~\ref{sec:formulation}. Then, we explain how to extract the multi-grained visual and textual features in Sec.~\ref{sec:extraction}. We show the details of the key bi-directional one-to-many embedding paradigm \module{} in Sec.~\ref{sec:embedding}. Finally, we describe the optimization of our proposed framework in Sec.~\ref{sec:optimization}.

\subsection{Problem Formulation}\label{sec:formulation}

The goal of the proposed \short{} is to calculate the semantic similarity scores between the description queries and the images in TPR. Formally, a TPR dataset consists of $P$ image-text pairs, which are denoted as $D = \left\{\hat{v}_i, \hat{t}_i\right\}|_{i=1}^P$. Each image-text pair $\left\{\hat{v}_i, \hat{t}_i\right\}$ is also provided with a identity (ID) label $y_i$. The set of all IDs is denoted as $Y=\{y_i\}|_{i=1}^P$ with $y_i \in\{1, \cdots, Q\}$, where $Q$ is the number of distinct IDs for all pedestrians. Given a textual description $\hat{t}_i$, a TPR model aims to  retrieve images with the same ID as $\hat{t}_i$ from a large-scale image gallery.

\subsection{Multi-Grained Feature Extraction}\label{sec:extraction}

As TRP is a fine-grained retrieval task, it is insufficient to extract only coarse-grained features to differentiate detailed information among pedestrians. To address this challenge, we follow previous research~\cite{ding2021semantically,shao2022learning} and extract multi-grained features from both images and texts to enable comprehensive perception. This approach allows us to capture more nuanced information and better distinguish between similar-looking pedestrians in the dataset.

\noindent \textbf{Backbone.}
Given an image $\hat{v}$, we use the pre-trained ResNet~\cite{he2016deep} backbone on ImageNet~\cite{deng2009imagenet} to extract the visual representation $\bm{V} \in \mathbb{R}^{H \times W \times C}$, where $H$, $W$ and $C$ are the height, width, and channel dimension of $\bm{V}$, respectively.

For a description sentence $\hat{t}$, we adopt a pre-trained frozen BERT~\cite{devlin2018bert} and a trainable Bi-LSTM model to extract the textual feature $\bm{T} \in \mathbb{R}^{N \times C}$ following \cite{wang2022caibc}, where $N$ and $C$ refer to the length and channel dimension of the textual representation. 

\noindent \textbf{Global Feature Extraction.}
To obtain global visual and textual features, we employ a global max pooling (GMP) operation on the visual feature map $\bm{V}$ and the textual feature map $\bm{T}$ to eliminate spatial dimensions. We then pass the resulting feature vectors through fully connected (FC) layers for embedding. This process can be expressed mathematically as follows:
\begin{equation} 
    \boldsymbol{v}_g = \mathrm{GMP}(\bm{V})\mathbf{W}_g^v,
\label{eq:global_v}
\end{equation}
\begin{equation} 
    \boldsymbol{t}_g =  \mathrm{GMP}(\bm{T}) \mathbf{W}_g^t,
\label{eq:global_t}
\end{equation}
where $\mathbf{W}_g^v \in \mathbb{R}^{C \times C_g} $ and $\mathbf{W}_g^t \in \mathbb{R}^{C \times C_g}$ are the trainable parameter matrices of FC layers. $C_g$ is the channel dimension of the global feature. $\boldsymbol{v}_g$ and $\boldsymbol{t}_g$ refer to the obtained global visual and textual features, respectively.

\noindent \textbf{Local Feature Extraction.}
To obtain local visual features, we first uniformly split $\bm{V}$ into $K$ non-overlapping parts, which is denoted as $\{\bm{\mathcal{V}}^k \in \mathbb{R}^{H^{\prime} \times W \times C}\}|_{k=1}^K $ with $H^{\prime} = \frac{H}{K}$. For textual features, we adopt the word attention module (WAM)~\cite{ding2021semantically} to obtain $K$ textual features $\{\bm{\mathcal{T}}^k \in \mathbb{R}^{N \times C}\}|_{k=1}^K$.

Then, for the $k$-th part features $\bm{\mathcal{V}}^k$ and $\bm{\mathcal{T}}^k$, we use GMP modules and FC layers to obtain the local visual feature $\boldsymbol{v}_l^k$ and textual feature $\boldsymbol{t}_l^k$:
\begin{equation} 
    \boldsymbol{v}_l^k = \mathrm{GMP}(\bm{\mathcal{V}}^k)\mathbf{W}_l^v,
\label{eq:local_v}
\end{equation}
\begin{equation} 
    \boldsymbol{t}_l^k = \mathrm{GMP}(\bm{\mathcal{T}}^k)\mathbf{W}_l^t,
\label{eq:local_t}
\end{equation}
where $\mathbf{W}_l^v, \mathbf{W}_l^t \in \mathbb{R}^{C \times C_l} $.

\noindent \textbf{Non-local Feature Extraction.}
For non-local interactions, we introduce the non-local module (NLM)~\cite{ding2021semantically} to perform modeling between local features to obtain non-local features. Specifically, we first calculate the cosine similarity score $s_{ki}$ between $\boldsymbol{v}_l^k$ and $\boldsymbol{v}_l^i$ after linear embedding:
\begin{equation} 
    s_{ki} = \frac{(\boldsymbol{v}_l^k \mathbf{W}_\alpha^k) \, (\boldsymbol{v}_l^i \mathbf{W}_\beta^i)^T}{\| \boldsymbol{v}_l^k \mathbf{W}_\alpha^k \| \,  \| \boldsymbol{v}_l^i \mathbf{W}_\beta^i\|},
\label{eq:cos}
\end{equation}
where $\mathbf{W}_\alpha^k, \mathbf{W}_\beta^i \in \mathbb{R}^{C_l \times C_l}$. Then, we aggregate local visual features based on similarity scores as follows:
\begin{equation} 
    \boldsymbol{v}_{agg}^k=\bigg(\sum_{i=1,i \neq k}^K \Big(\frac{\exp(s_{ki})}{\sum_{j=1,j \neq k}^K\exp(s_{kj})} \big(\boldsymbol{v}_l^i \mathbf{W}_\beta^i \big) \Big) \bigg) \mathbf{W}_\gamma^k,
\label{eq:agg}
\end{equation}
where $\mathbf{W}_\gamma^k \in \mathbb{R}^{C_l \times C_l}$. The final non-local visual feature is denoted as follows:
\begin{equation} 
    \boldsymbol{v}_{n}^k=(\boldsymbol{v}_l^k+\boldsymbol{v}_{agg}^k)\mathbf{W}_\delta^k,
\label{eq:non}
\end{equation}
where $\mathbf{W}_\delta^k \in \mathbb{R}^{C_l \times C_n}$. The non-local textual feature could be obtained in the same way.

\subsection{Bi-directional One-to-Many
 Embedding}\label{sec:embedding}

The conventional modal-shared one-to-one embedding paradigm~\cite{zhu2021dssl,zheng2020hierarchical,ding2021semantically,chen2018improving} involves embedding visual and textual samples into a joint space for alignment using a one-to-one projection function. However, this approach overlooks the challenges posed by uncertain optimization and the one-to-many relationships present in image-text pairs. To overcome these limitations, we propose a novel bi-directional one-to-many embedding paradigm. In this section, we provide detailed explanations of both the bi-directional embedding and one-to-many embedding techniques that we employ in our proposed approach.

\noindent \textbf{Bi-directional Embedding.}
As analyzed in Sec.~\ref{sec:intro}, since features of both modalities are movable in the joint space, the conventional modal-shared projection leads to the problem of uncertain optimization direction, thus increasing the difficulty of optimization. 
To address the challenges posed by the movable features in the joint space, we propose a bi-directional embedding paradigm that projects the features of each modality into their respective modal-specific spaces. Specifically, we first fix the textual features and project only the visual features into the textual feature space for matching. This approach provides the visual features with a clear optimization direction. We also fix the visual features and only project textual features into the image space for alignment. Formally, the bi-directional embedding for global features can be expressed as follows:
\begin{equation} 
    \boldsymbol{v}_{g \_ txt} = f^{v2t}(\boldsymbol{v}_g),
\label{eq:bi_v}
\end{equation}
\begin{equation} 
    \boldsymbol{t}_{g \_ img} = f^{t2v}(\boldsymbol{t}_g),
\label{eq:bi_t}
\end{equation}
where $\boldsymbol{v}_{g\_txt}$ and $\boldsymbol{t}_{g\_img}$ represent the global visual feature in text space and the global textual feature in image space, respectively. To project visual features into text space, we employ an embedding function denoted by $f^{v2t}(\cdot)$, while for projecting textual features into image space, we use the embedding function $f^{t2v}(\cdot)$. The next section provides detailed information about the specific forms of these embedding functions.

 
\noindent \textbf{One-to-Many Embedding.}
The one-to-many relationship between image and text pairs is a widely recognized fact, and the conventional one-to-one projection may yield suboptimal outcomes in TPR datasets. The reason behind this is that the multiple positive samples of a query may have slight variations. If only the one-to-one projection is employed on query samples, it becomes challenging for multiple positive samples to align with the unique embedding feature of the query. To address this issue and account for the one-to-many relationship in image-text pairs, we propose a residual embedding module group (\module{}) as $f^{v2t}(\cdot)$ and $f^{t2v}(\cdot)$. This module group is capable of performing one-to-many embedding.
Our proposed \module{} consists of $M$ residual embedding modules (REMs). Each REM incorporates a residual-based linear embedding. The $m$-th REM for global features is formulated as follows:
\begin{equation} 
    \boldsymbol{v}_{g \_ txt}^m = \boldsymbol{v}_g + \sigma(\boldsymbol{v}_g \mathbf{W}_{ v2t}^1)\mathbf{W}_{ v2t}^2,
\label{eq:bi_v2}
\end{equation}
\begin{equation} 
    \boldsymbol{t}_{g \_ img}^m = \boldsymbol{t}_g + \sigma(\boldsymbol{t}_g \mathbf{W}_{t2v}^1)\mathbf{W}_{t2v}^2,
\label{eq:bi_t2}
\end{equation}
where $\mathbf{W}_{ v2t}^1, \mathbf{W}_{t2v}^1 \in \mathbb{R}^{C_g \times \frac{C_g}{r}}$ and $\mathbf{W}_{ v2t}^2, \mathbf{W}_{t2v}^2 \in \mathbb{R}^{\frac{C_g}{r} \times C_g}$ are trainable parameter matrices of the $m$-th REM. $\sigma(\cdot)$ is ReLU activation function. $\boldsymbol{v}_{g \_ txt}^m$ and $\boldsymbol{t}_{g \_ img}^m$ are the global visual features in the text space and the global textual features in the image space, which are projected by the $m$-th REM. Since REM-G consists of $M$ REMs, the outputs of REM-G are a list of $M$ visual and textual features, \emph{i.e.,} $ \boldsymbol{v}_{g \_ txt }=\{\boldsymbol{v}_{g \_ txt}^m\}|_{m=1}^M$ and  $\boldsymbol{t}_{g \_ img} = \{\boldsymbol{t}_{g \_ img}^m\}|_{m=1}^M$.

Similarly, following the bi-directional one-to-many embedding paradigm, local and non-local features also yield multiple embedding features in both text and image spaces.


\subsection{Optimization}\label{sec:optimization}
To optimize the model effectively, we employ two widely used objectives, namely, the identity (ID) loss and the compound ranking (CR) loss, which have been previously used in \cite{ding2021semantically}.

\noindent \textbf{ID Loss.} We regard pedestrians with different IDs as different classes and cluster images into groups according to their IDs via ID loss. ID loss is formulated as follows:
\begin{equation} 
\begin{aligned}
    \mathcal{L}_{I D}(\boldsymbol{v},\boldsymbol{t},\boldsymbol{y})=-\log \left(\operatorname{Softmax}\left(\boldsymbol{v} \mathbf{W}^v_{i d}\right)[\boldsymbol{y}]\right)\\
    -\log \left(\operatorname{Softmax}\left(\boldsymbol{t} \mathbf{W}^t_{i d}\right)[\boldsymbol{y}]\right),
\label{eq:id}
\end{aligned}
\end{equation}
where $\boldsymbol{v}$, $\boldsymbol{t}$, and $\boldsymbol{y}$ are visual features, textual features, and the corresponding IDs, respectively. $\mathbf{W}^v_{id},\mathbf{W}^t_{id} \in \mathbb{R}^{C_{in} \times Q}$ are trainable parameter matrices. $[\cdot]$ is the operation that takes the value according to the index. $C_{in}$ is the channel dimension of input features and $Q$ is the number of person IDs in the training set. 

\noindent \textbf{CR Loss.} 
The ranking-based loss has shown its strength in image-text matching ~\cite{lee2018stacked,faghri2017vse++}. Following \cite{ding2021semantically}, we adopt CR loss as a matching objective:
\begin{equation} 
    \begin{aligned}
\mathcal{L}_{CR} &=\max \left(\alpha_1-S\left(\boldsymbol{v}, \boldsymbol{t}\right)+S(\boldsymbol{v}, \hat{\boldsymbol{t}}), 0\right) \\
&+\max \left(\alpha_1-S\left(\boldsymbol{v}, \boldsymbol{t}\right)+S\left(\hat{\boldsymbol{v}}, \boldsymbol{t}\right), 0\right) \\
&+\beta \cdot \max (\alpha_2-S\left(\boldsymbol{v}, \boldsymbol{t}^{\prime}\right)+S(\boldsymbol{v}, \hat{\boldsymbol{t}^\prime}), 0) \\
&+\beta \cdot \max \left(\alpha_2-S\left(\boldsymbol{v}, \boldsymbol{t}^{\prime}\right)+S\left(\hat{\boldsymbol{v}}, \boldsymbol{t}^{\prime}\right), 0\right),
\end{aligned}
\label{eq:cr}
\end{equation}
where $\alpha_2 = \left(\min \left(
\frac{S(\boldsymbol{v}, \boldsymbol{t}^{\prime})}{S(\boldsymbol{v},\boldsymbol{t})}
, 1\right)+1\right) \frac{\alpha_1}{2}$. $\left(\boldsymbol{v}, \boldsymbol{t}\right)$ and $\left(\boldsymbol{v}, \boldsymbol{t}^{\prime}\right)$ represent matched image-text pairs. $(\boldsymbol{v}, \hat{\boldsymbol{t}})$, $\left(\hat{\boldsymbol{v}}, \boldsymbol{t}\right)$, $\left(\boldsymbol{v}, \hat{\boldsymbol{t}^\prime}\right)$ and $\left(\hat{\boldsymbol{v}}, \boldsymbol{t}^{\prime}\right)$ are unmatched pairs. $\boldsymbol{t}^{\prime}$ is the text for another image with the same identity as $\boldsymbol{v}$. $\alpha_1$ and $\alpha_2$ refer to the margins and $\beta$ indicates the weight for the weak supervision terms. 

In the conventional methods, there is only one feature for each sample in the modal-shared space, so $S(\cdot,\cdot)$ is just the cosine similarity between the two samples. However, in our method, there are two modal-specific spaces for alignment, and each sample produces $M$ features after projection. To this end, we sum the similarity scores computed on multiple embedded features in these two spaces as the final similarity score, which is formulated as:
\begin{equation} 
    \begin{aligned}
S(\boldsymbol{v}, \boldsymbol{t}) = \sum _{m=1}^M s(\boldsymbol{v}, \boldsymbol{t}^m_{img}) + \sum _{m=1}^M s(\boldsymbol{v}^m_{txt}, \boldsymbol{t}),
\end{aligned}
\label{eq:simi}
\end{equation}
where $M$ is the number of REMs in \module{}. $\boldsymbol{t}^m_{img}$ and $\boldsymbol{v}^m_{txt}$ are the
visual feature in the text space and the textual feature in the image space projected by the $m$-th REM, $s(\cdot,\cdot)$ is the cosine similarity function. $\boldsymbol{v}$ and $\boldsymbol{t}$ include global, local and non-local features.

The final loss function is the sum of ID loss and CR loss:
\begin{equation} 
    \begin{aligned}
\mathcal{L} = \mathcal{L}_{ID} + \mathcal{L}_{CR}.
\end{aligned}
\label{eq:loss}
\end{equation}

\section{Experiment}

\subsection{Datasets}
To evaluate the effectiveness of \short{}, we conducted experiments on three widely-used TRP datasets: \emph{CUHK-PEDES}~\cite{li2017person}, \emph{ICFG-PEDES}~\cite{ding2021semantically}, and \emph{RSTPReID}~\cite{zhu2021dssl}. In addition, to verify the generality of \short{}, we also evaluated its performance of three other image-text retrieval datasets: \emph{MS-COCO}~\cite{lin2014microsoft}, \emph{Caltech-UCSD Birds (CUB)}~\cite{reed2016learning}, and \emph{Oxford-102 Flowers (Flowers)}~\cite{reed2016learning}. Further details on these datasets are available in the supplementary materials.

\begin{table}[]
\caption{Comparison with SOTA methods on CUHK-PEDES. * indicates that we use ResNet-101 as the backbone.}
\vspace{-0.3cm}
\centering
\setlength{\tabcolsep}{4.5mm}
\begin{tabular}{l|ccc}
\Xhline{1.0pt}
Model    & R@1            & R@5            & R@10           \\
\hline
CNN-RNN~\cite{reed2016learning}  & 8.07           & -              & 32.47          \\
GNA-RNN~\cite{li2017person}  & 19.05          & -              & 53.64          \\
PWM-ATH~\cite{chen2018improving}  & 27.14          & 49.45          & 61.02          \\
GLA~\cite{chen2018improving}      & 43.58          & 66.93          & 76.26          \\
MIA~\cite{niu2020improving}      & 53.10          & 75.00          & 82.90          \\
A-GANet~\cite{liu2019deep}  & 53.14          & 74.03          & 81.95          \\
ViTAA~\cite{wang2020vitaa}    & 55.97          & 75.84          & 83.52          \\
IMG-Net~\cite{wang2020img}  & 56.48          & 76.89          & 85.01          \\
CMAAM~\cite{aggarwal2020text}    & 56.68          & 77.18          & 84.86          \\
HGAN~\cite{zheng2020hierarchical}     & 59.00          & 79.49          & 86.60          \\
DSSL~\cite{zhu2021dssl}     & 59.98          & 80.41          & 87.56          \\
MGEL~\cite{wang2021text}     & 60.27          & 80.01          & 86.74          \\
SSAN~\cite{ding2021semantically}     & 61.37          & 80.15          & 86.73          \\
NAFS~\cite{gao2021contextual}     & 61.50          & 81.19          & 87.51          \\
TBPS~\cite{han2021text}     & 61.65          & 80.98          & 86.78          \\
LapsCore~\cite{wu2021lapscore} & 63.40          & -              & 87.80          \\
\hline
\short{}     & 64.23 & 82.91 & 88.65 \\
\short{}* & \textbf{65.61} & \textbf{83.45} & \textbf{89.57}\\
\Xhline{1.0pt}
\end{tabular}
\label{tab:cuhk_performance}
\end{table}

\subsection{Experimental Settings}

We conduct the experiments on RTX3090 24GB GPUs using the PyTorch library. We set the maximum sentence length to 100 and resize the input images to $384 \times 112$ with random horizontal flipping augmentation. We empirically set $M$ to 4 and adopt the values of $K, C, C_g, C_l, C_n, \alpha_1, \beta$ from previous work \cite{ding2021semantically, wang2020vitaa}, specifically 6, 2048, 1024, 1024, 512, 0.2, and 0.1, respectively. $r$ is set to 8. We optimize the \short{} model for 80 epochs using the Adam optimizer \cite{kingma2014adam} with a batch size of 64 and a learning rate of 0.001. We weight the loss functions for global, local, and non-local features as 2, 1, and 1, respectively. During inference, we calculate the cosine similarity between the query feature and $M$ candidate features in each space, and select the maximum value as the similarity in this space. The final similarity is the sum of the similarities calculated in the two spaces. In the following experimental results (\emph{i.e.,} Tab.~\ref{tab:speed} and Fig.~\ref{fig:heatmap}, \ref{fig:cub}), the only difference between the base model and \short{} is that the base model uses the modal-shared one-to-one embedding paradigm \cite{ding2021semantically} using the plain linear layer, while \short{} adopts a bi-directional one-to-many embedding paradigm.

\noindent \subsection{Performance Comparison}


To demonstrate the effectiveness and generalizability of our proposed method, we conducted extensive experiments and compared it against previous state-of-the-art (SOTA) models on three widely-used benchmarks: CUHK-PEDES, ICFG-PEDES, and RSTPReID. Our results, as shown in Tab.~\ref{tab:cuhk_performance}-\ref{tab:rstp_performance}, consistently demonstrate that \short{} outperforms the strongest competitors by a significant margin. Specifically, we achieved 65.61 R@1, 58.25 R@1, and 48.10 R@1 on these datasets, respectively. It is noteworthy that previous TPR methods only produce a single feature for each sample and calculate cross-modal similarity score in the modal-shared space, whereas our proposed \short{} is a pioneer work that explores the bi-directional one-to-many embedding paradigm for cross-modal alignment. Our significant performance gain clearly shows that this new paradigm is highly effective in aligning multi-modal samples and improving the TPR task's overall performance.

\begin{table}[]
\caption{Comparison with SOTA methods on ICFG-PEDES.}
\vspace{-0.3cm}
\centering
\setlength{\tabcolsep}{4.0mm}{
\begin{tabular}{l|ccc}
\Xhline{1.0pt}
Model     & R@1            & R@5            & R@10           \\
\hline
Dual Path~\cite{zheng2020dual} & 38.99          & 59.44          & 68.41          \\
CMPM+CMPC~\cite{zhang2018deep} & 43.51          & 65.44          & 74.26          \\
MIA~\cite{niu2020improving}       & 46.49          & 67.14          & 75.18          \\
SCAN~\cite{lee2018stacked}      & 50.05          & 69.65          & 77.21          \\
ViTAA~\cite{wang2020vitaa}     & 50.98          & 68.79          & 75.78          \\
SSAN~\cite{ding2021semantically}      & 54.23          & 72.63          & 79.53          \\
\hline
\short{}      & 57.62 & 75.04 & 81.53 \\
\short{}* & \textbf{58.25} & \textbf{75.92} & \textbf{81.96} \\
\Xhline{1.0pt}
\end{tabular}
}
\label{tab:icfg_performance}
\vspace{-0.3cm}
\end{table}

\begin{figure}
\centering 
  \includegraphics[width=0.9\columnwidth]{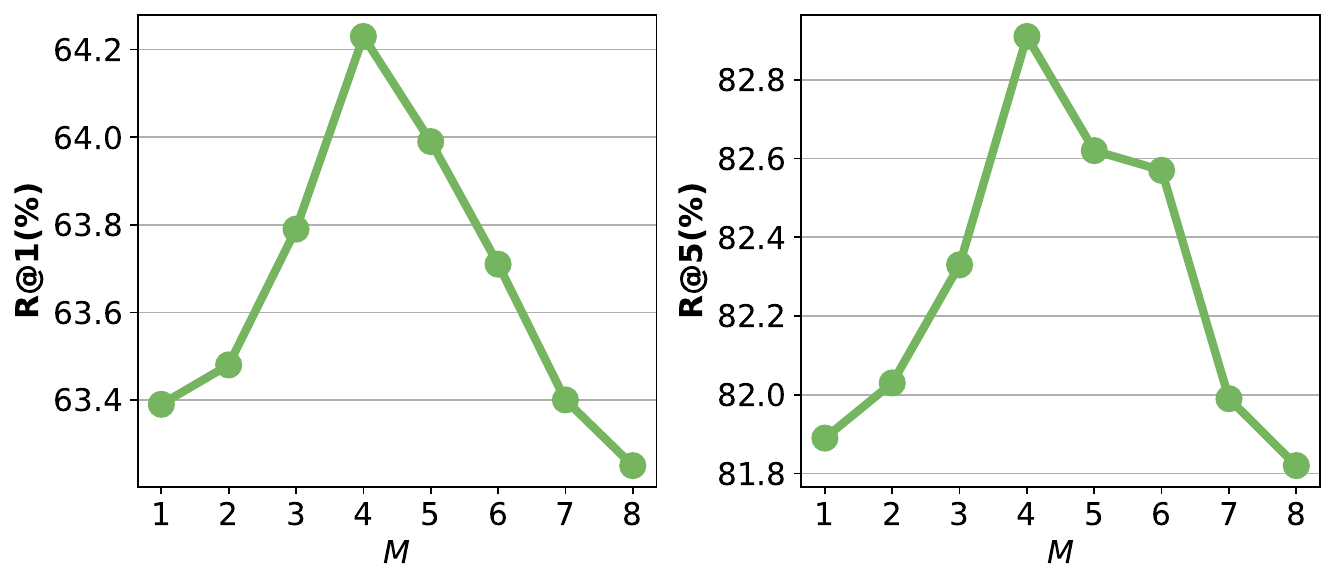}
  \vspace{-0.4cm}
  \caption{ Effect of different numbers of REMs (\emph{i.e.,} $M$) in each REM-G on the CUHK-PEDES dataset.
}
  \label{fig:m}
\end{figure}

\begin{table}[]
\vspace{-0.3cm}
\caption{Comparison with SOTA methods on RSTPReID.}
\vspace{-0.3cm}
\centering
\setlength{\tabcolsep}{4.5mm}{
\begin{tabular}{l|ccc}
\Xhline{1.0pt}
Model                                              & R@1            & R@5            & R@10           \\
\hline
IMG-Net~\cite{wang2020img}   & 37.60          & 61.15          & 73.55          \\
AMEN~\cite{wang2021amen}     & 38.45          & 62.40          & 73.80          \\
DSSL~\cite{zhu2021dssl}      & 32.43          & 55.08          & 63.19          \\
ASPD-Net~\cite{wang2022aspd} & 39.90          & 65.15          & 74.40          \\
SUM~\cite{wang2022sum}       & 41.38          & 67.48          & 76.48          \\
\hline
\short{}   & 47.30 & 69.65 & 79.20 \\
\short{}*  & \textbf{48.10} & \textbf{73.10} & \textbf{81.30} \\
\Xhline{1.0pt}
\end{tabular}
}
\label{tab:rstp_performance}
\vspace{-0.4cm}
\end{table}

\subsection{Ablation Study}

\begin{table}[]
\caption{ Performance and efficiency analysis of the Base model and \short{}. All experiments are conducted on a single NVIDIA GeForce RTX 3090 GPU.}
\vspace{-0.3cm}
\centering
\setlength{\tabcolsep}{4.5mm}
\begin{tabular}{l|ccc}
\Xhline{1pt}
Model                     & Params  & Speed       & R@1   \\ \hline
Base                      & 230.13M & 31.38 ms/txt & 62.78  \\
\short{} & 230.13M & 34.93 ms/txt & 64.23 \\ \Xhline{1pt}
\end{tabular}
\vspace{-0.4cm}
\label{tab:speed}
\end{table}

\noindent \textbf{Efficiency and Parameter Comparison.} Tab.~\ref{tab:speed} presents a comparison of the efficiency and parameters between the base model and  \short{}. By adopting a bottleneck structure, \short{} maintains the same number of parameters as the base model. Thus, the performance gain does not come from the increase of trainable parameters.  Due to the one-to-many embedding of \short{}, cross-modal similarity calculations are performed multiple times, which may affect the model efficiency. Nevertheless, our experimental results demonstrate that the impact on efficiency (+3.55 ms/txt) can be disregarded, given the significant performance improvements achieved (+1.45 R@1).


\noindent \textbf{Effect of Embedding Space.} To gain a deeper understanding of the bi-directional embedding paradigm, we conducted a series of ablation studies on the CUHK-PEDES dataset, in which we aligned image and text samples in different spaces, namely the image space, text space, and modal-shared space. Our analysis of the results presented in Tab.~\ref{tab:space} led us to the following observations:
\begin{itemize}[leftmargin=*, itemsep=2pt,topsep=0pt,parsep=0pt]    
    \item We found that aligning samples in both image and text spaces (Exp7) outperforms aligning them in either space alone (Exp4, Exp5) by a margin of 1.72 R@1 and 2.05 R@1, respectively. This indicates that the alignment results in the image and text spaces can complement each other and lead to better retrieval performance, thereby demonstrating the superiority of our proposed bi-directional embedding paradigm.

    \item To eliminate the effect of the number of alignment spaces, we further conducted experiments by aligning samples in different numbers of modal-shared spaces (Exp1, Exp2, Exp3). Our analysis reveals that the retrieval performance declines slightly as the number of modal-shared spaces (MSS) increases. This finding indicates that the performance improvement of our bi-directional embedding paradigm is not solely attributable to the increase in the number of alignment spaces, but rather to the effectiveness of the paradigm itself.

    \item To further explore the impact of alignment in MSS, we conducted an experiment (Exp6) by aligning visual and textual samples in all three spaces. However, we found that the performance of Exp6 was worse than that of Exp7, suggesting that alignment in MSS leads to performance degradation. The main reason for this may be that alignment in MSS results in the problem of uncertain optimization direction, as explained in Sec.~\ref{sec:intro}.

    \item By comparing Exp1, Exp4, and Exp5, we observed that aligning samples in the text space performs slightly worse than aligning them in either the image or modal-shared space. This could be because the expression of text is often abstract and general, making it more challenging to align samples in the text space compared to the image space and modal-shared space.
    
\end{itemize}

\begin{table}[]
\caption{ Ablation studies to explore the effect of embedding space on CUHK-PEDES. ``MSS'', ``IS'' and ``TS'' are short for ``Modal-Shared Space'', ``Image Space'' and ``Text Space'', respectively. The number of $\checkmark$ denotes the number of spaces.}
\vspace{-0.3cm}
\centering
\setlength{\tabcolsep}{2.5mm}{
\begin{tabular}{l|l|l|l|ccc}
\Xhline{1.0pt}
ID  & MSS                                                                           & IS                        & TS                        & R@1             & R@5             & R@10            \\
\hline
Exp1 & \checkmark                                                     &                           &                           & 62.78          & 81.43          & 87.52          \\
Exp2 & \checkmark \checkmark                           &                           &                           & 62.46          & 80.99          & 87.64          \\
Exp3 & \checkmark \checkmark \checkmark &                           &                           & 62.43          & 80.45          & 87.46          \\
Exp4 &                                                                               & \checkmark &                           & 62.51          & 82.02          & 87.85          \\
Exp5 &                                                                               &                           & \checkmark & 62.18          & 81.22          & 86.81          \\
Exp6 &  \checkmark &   \checkmark  & \checkmark & 63.01& 	81.60& 	87.82          \\
\hline
Exp7 &                                                                               & \checkmark & \checkmark & \textbf{64.23} & \textbf{82.91} & \textbf{88.65} \\
\Xhline{1.0pt}
\end{tabular}
}
\vspace{-0.2cm}
\label{tab:space}
\end{table}

\begin{table}[]
\caption{ Ablation of loss function on CUHK-PEDES. }
\vspace{-0.3cm}
\centering
\setlength{\tabcolsep}{3.5mm}{
\begin{tabular}{ll|ccc}
\Xhline{1.0pt}
ID Loss & CR Loss & R@1            & R@5            & R@10           \\
\hline
\checkmark        &            & 47.43          & 70.86          & 80.04          \\
         & \checkmark          & 37.23          & 59.47          & 69.57          \\
\checkmark        & \checkmark          & \textbf{64.23} & \textbf{82.91} & \textbf{88.65} \\
\Xhline{1.0pt}
\end{tabular}
}
\label{tab:loss}
\vspace{-0.2cm}
\end{table}


\noindent \textbf{Effect of Loss Function.} To explore the impact of the loss function, we also conduct ablation studies by adopting different loss functions. \short{} is equipped with two loss functions, \emph{i.e.,} ID loss and CR loss, to both minimize the intra-class distance and maximize the inter-class distance. As shown in Tab. \ref{tab:loss}, we observe that both ID loss and CR loss are critical for retrieval performance. In particular, the lack of CR loss and ID loss leads to significant performance degradation of 16.80 R@1 and 27.00 R@1, respectively.

\begin{figure*}
\centering 
  \includegraphics[width=1.8\columnwidth]{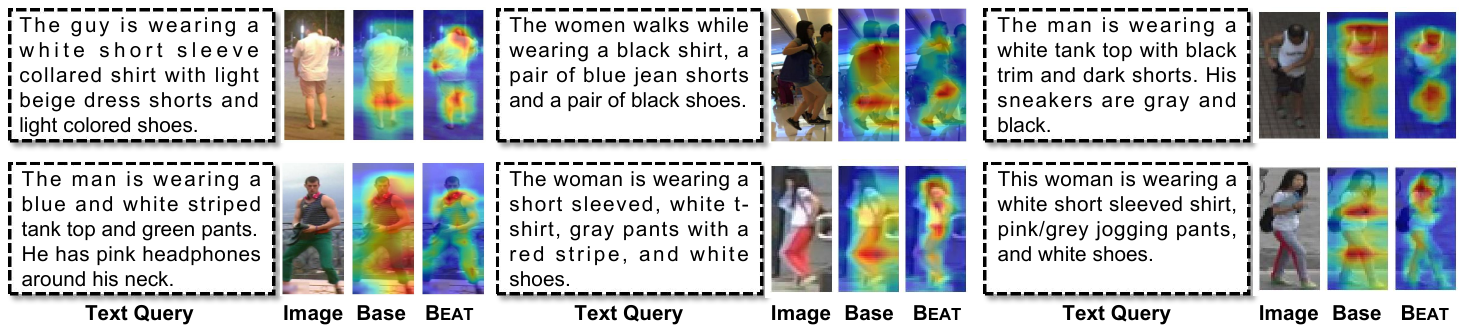}
  \vspace{-0.4cm}
  \caption{ Comparison of heatmaps between the base model and \short{}. The Base model often attends to a large area, while the BEAT model can accurately focus on pedestrian areas.
  }
\vspace{-0.3cm}
  \label{fig:heatmap}
\end{figure*}


\begin{table}[]
\caption{ Ablation of multi-grained features on CUHK-PEDES.}
\vspace{-0.3cm}
\centering
\resizebox{1.00\columnwidth}{!}{
\begin{tabular}{lll|ccc}
\Xhline{1.0pt}
Local & Non-local & Global & R@1            & R@5            & R@10           \\
\hline
\checkmark     &           &        & 59.31          & 79.48          & 86.57          \\
      & \checkmark         &        & 61.37          & 80.80           & 87.22          \\
      &           & \checkmark      & 57.81          & 77.45          & 84.34          \\
      \hline
\checkmark     & \checkmark         &        & 61.61          & 81.12          & 87.31          \\
\checkmark     &           & \checkmark      & 61.92          & 81.14          & 87.91          \\
      & \checkmark         & \checkmark      & 61.84          & 80.88          & 87.59          \\
      \hline
\checkmark     & \checkmark         & \checkmark      & \textbf{64.23} & \textbf{82.91} & \textbf{88.65} \\
\Xhline{1.0pt}
\end{tabular}
}
\label{tab:grained}
\end{table}

\begin{table}[]
\caption{  Quantitative evaluation results on MS-COCO 1K test set. The best results are in bold. We report the performance of a single model without any ensemble technology.}
\vspace{-0.3cm}
\centering
\resizebox{1.00\columnwidth}{!}{
\begin{tabular}{l|ccc|ccc|c}
\Xhline{1.0pt}
\multirow{2}{*}{Model} & \multicolumn{3}{c}{Image-to-Text}             & \multicolumn{3}{|c|}{Text-to-Image}             & \multirow{2}{*}{rSum} \\ \cline{2-7}
                       & R@1           & R@5           & R@10          & R@1           & R@5           & R@10          &                       \\ \hline
SCAN~\cite{lee2018stacked}                   & 72.7          & 94.8          & 98.4          & 58.8          & 88.4          & 94.8          & 507.9                 \\
BFAN~\cite{liu2019focus}                   & 74.9          & 95.2          & -             & 59.4          & 88.4          & -             & 317.9                 \\
PVSE~\cite{song2019polysemous}                   & 69.2          & 91.6          & 96.6          & 55.2          & 86.5          & 93.7          & 492.8                 \\
CVSE~\cite{wang2020consensus}                   & 69.2          & 93.3          & 97.5          & 55.7          & 86.9          & 93.8          & 496.4                 \\
DPRNN~\cite{chen2020expressing}                  & 75.3          & 95.8          & 98.6          & 62.5          & 89.7          & 95.1          & 517.0                 \\
SGM~\cite{wang2020cross}                    & 73.4          & 93.8          & 97.8          & 57.5          & 87.3          & 94.3          & 504.1                 \\
IMRAM~\cite{chen2020imram}                  & 76.7          & 95.6          & 98.5          & 61.7          & 89.1          & 95.0          & 516.6                 \\
SMFEA~\cite{ge2021structured}                  & 75.1          & 95.4          & 98.3          & 62.5          & 90.1          & 96.2          & 517.6                 \\
SHAN~\cite{ji2021step}                   & 76.8          & 95.4          & 98.7          & 62.6          & 89.6          & 95.8          & 519.8                 \\
NAAF~\cite{zhang2022negative}                   & 78.1          & 96.1          & 98.6          & 63.5          & 89.6          & 95.3          & 521.2                 \\ \hline
NAAF + \module{}               & \textbf{79.0} & \textbf{96.9} & \textbf{98.9} & \textbf{64.1} & \textbf{90.2} & \textbf{95.5} & \textbf{524.7}       \\ 
\Xhline{1.0pt}
\end{tabular}
}
\label{tab:coco}
\vspace{-0.3cm}
\end{table}

\noindent \textbf{Effect of Multi-grained Features.} To explore the impact of multi-grained features, we conduct ablation studies incrementally. As reported in Tab.~\ref{tab:grained}, the retrieval performance will gradually improve as more multi-grained features are adopted. These facts demonstrate that multi-grained features may provide complementary information to represent images and descriptions. Besides, we observe that if only one type of feature is used, non-local features (61.37 R@1) outperform global (57.81 R@1) and local (59.31 R@1) features significantly. This may be because non-local features not only contain part-level information but also have global information through non-local interaction.

\noindent \textbf{Effect of $M$.}
We propose \module{} to implement one-to-many embedding and each \module{} consists of $M$ REMs, so $M$ determines the number of embedded features for each sample. To explore the impact of $M$ (\emph{i.e.,} the number of REMs in each \module{}), we conduct experiments by increasing it from 1 to 8. As shown in Fig.~\ref{fig:m}, we observe that increasing $M$ in an appropriate range (\emph{i.e.,} from 1 to 4) can improve the retrieval performance significantly. This can be because the proposed one-to-many embedding strategy handles the scenario of the one-to-many relationship in image-text pairs. However, when $M$ is greater than 4, performance begins to drop. The reason may be that too many embedded features increase the difficulty of optimization.

\subsection{Qualitative Analysis}

\noindent \textbf{Heat map.} To provide a deeper understanding of the visual and textual alignment, we have presented heat maps in Fig.~\ref{fig:heatmap}. These heat maps illustrate the similarity scores between the global textual feature $\boldsymbol{t}_g$ and all grids of the visual feature map $\bm{V}$, where the brighter parts indicate higher similarity scores. Notably, the description queries in both models mainly focus on the person's body. It is worth noting that the proposed \short{} model accurately localizes the person, even capturing their outline. On the other hand, the base model can only locate the general area of the person (as seen in row 1, column 2 of Fig.~\ref{fig:heatmap}). Thus, owing to the bi-directional one-to-many embedding paradigm, \short{} achieves stronger cross-modal semantic alignment capability compared to the base model.

\begin{figure*}
\centering 
  \includegraphics[width=1.9\columnwidth]{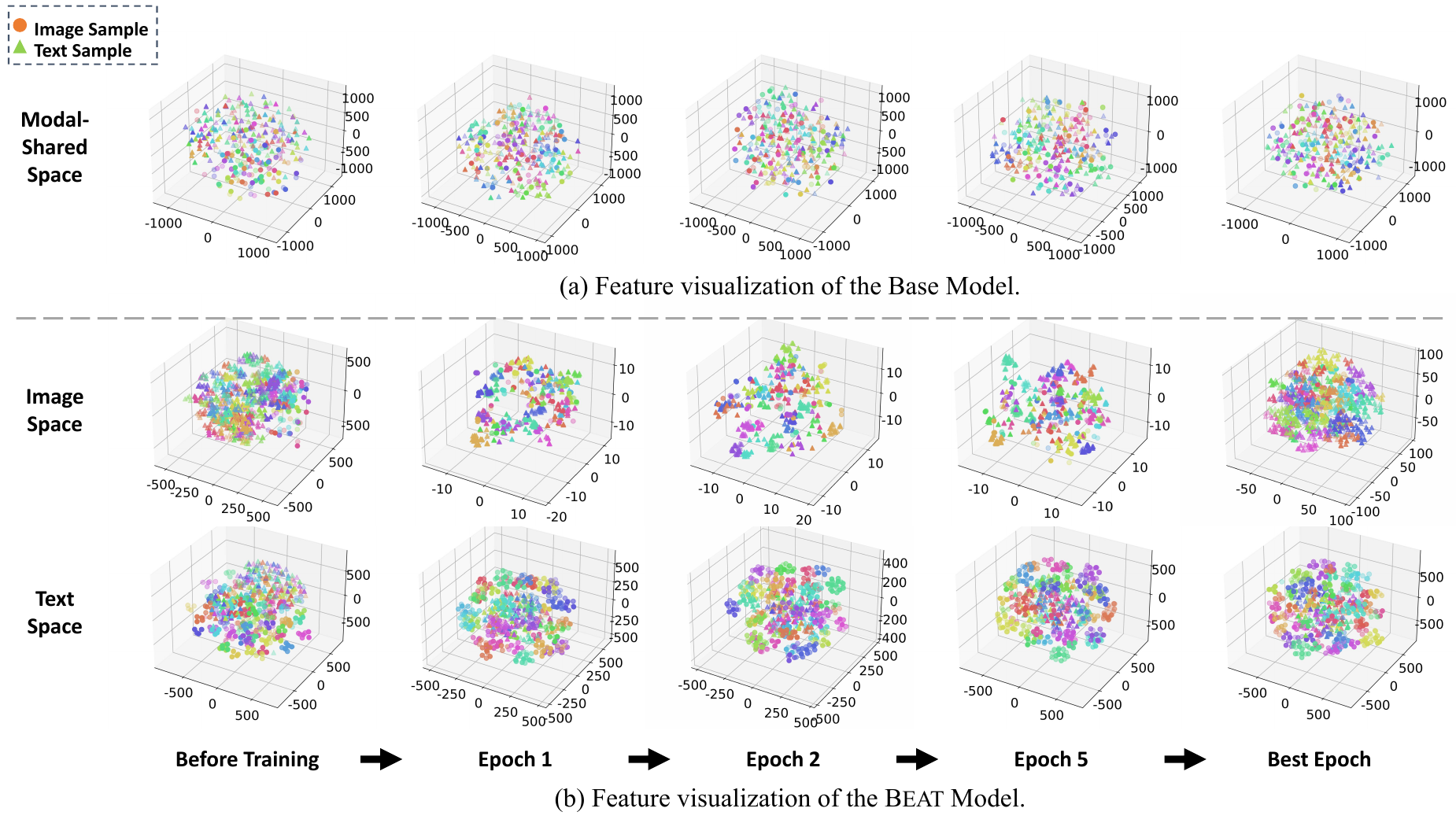}
  \vspace{-0.4cm}
  \caption{ Feature visualization of the base model and \short{} via t-SNE~\cite{van2008visualizing} on CUHK-PEDES. We show the changing process of cross-modal feature distributions with training. The feature of each image and text is marked as a circle and a triangle, respectively. Each identity is indicated in a specific color. For clarity, we only visualize global visual and textual features.
  }
  \label{fig:tSNE}
    \vspace{-0.2cm}
\end{figure*}

\noindent \textbf{t-SNE Visualization.}
We randomly sample the images and descriptions of 30 classes and visualize the visual and textual features of different training epochs via t-SNE~\cite{van2008visualizing} in Fig.~\ref{fig:tSNE} and obtain the following observations:
\begin{itemize}[leftmargin=*, itemsep=2pt,topsep=0pt,parsep=0pt]

    \item Before training, a significant gap exists between visual and textual modalities, and the distance between image and text samples of the same identity (ID) is large in either image or text space. After several training epochs, samples with the same ID are clustered together, and the modality gap is gradually eliminated. Finally, the samples with the same ID converge to a center, and the modality gap almost disappears.


    \item Compared with the base model, textual and visual features with the same ID generated by \short{} achieve better clustering results, which can be observed in the last column of Fig.\ref{fig:tSNE}.
    
\end{itemize}


\begin{figure}
\centering 
  \includegraphics[width=0.9\columnwidth]{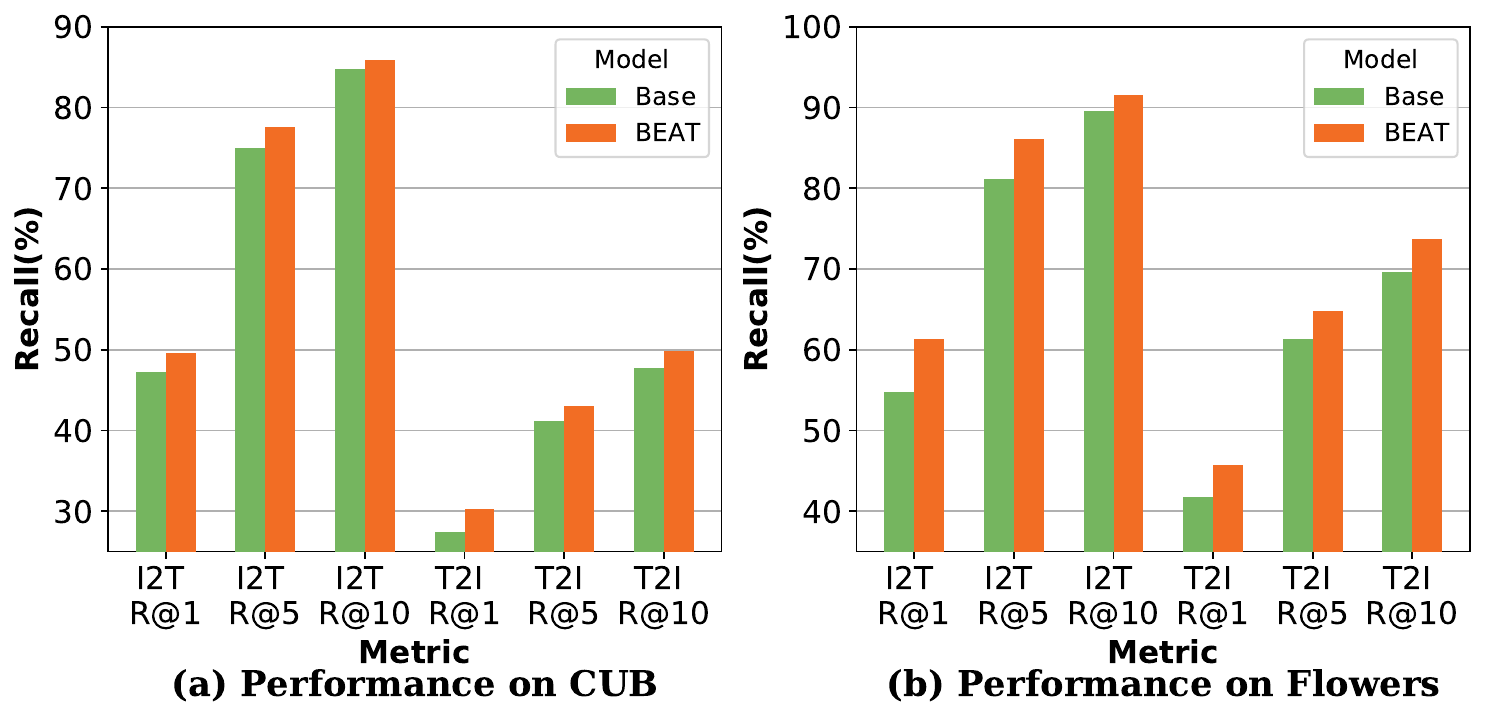}
  \vspace{-0.5cm}
  \caption{Performance comparisons between the base model and \short{} on the CUB and Flowers datasets.
  }
  \vspace{-0.4cm}
  \label{fig:cub}
\end{figure}

\section{ generalizability verification}
To verify the generalization of our method, we also conduct experiments on several image-text retrieval datasets, including MS-COCO, CUB, and Flowers.  Tab.~\ref{tab:coco} lists the SOTA performance comparison on MS-COCO. We equip the proposed \module{} on the previous SOTA model, \emph{i.e.,} NAAF~\cite{zhang2022negative}. It can be observed that NAAF+\module{} is +0.9 R@1 (I2T) and +0.6 R@1 (T2I) higher than NAAF. Furthermore, we conduct experiments on CUB~\cite{reed2016learning} and Flowers~\cite{reed2016learning} to compare the base model and \short{}. As shown in Fig.~\ref{fig:cub}, a consistent performance gain is also achieved via adopting the bi-directional one-to-many embedding paradigm, which demonstrates that the proposed paradigm is effective and universal for several types of multi-modal retrieval tasks.

\section{Conclusion}
This paper presents a novel \short{} model, which is the first work on exploring the bi-directional one-to-many embedding paradigm for TPR. \short{} achieves state-of-the-art performance on several text-based person retrieval datasets, including CUHK-PEDES, ICFG-PEDES, and RSTPReID. Besides, the proposed bi-directional one-to-many embedding paradigm also attains consistent performance improvement on several image-text retrieval datasets, \emph{i.e.,} MS-COCO, CUB, and Flowers. In the future, we will look forward to investigating its usage on uni-modal retrieval (\emph{e.g.,} image retrieval) and other multi-modal tasks (\emph{e.g.,} image captioning).

\section*{Acknowledgement}
This work was supported by National Key R\&D Program of China (No.2022ZD0118201), the National Science Fund for Distinguished Young Scholars (No.62025603), the National Natural Science Foundation of China (No. U21B2037, No. U22B2051, No. 62176222, No. 62176223, No. 62176226, No. 62072386, No. 62072387, No. 62072389, No. 62002305 and No. 62272401), China Postdoctoral Science Foundation (No.2023M732948), and the Natural Science Foundation of Fujian Province of China (No.2021J01002,  No.2022J06001).

\bibliographystyle{ACM-Reference-Format}
\balance
\bibliography{sample-base}


\begin{thebibliography}{64}


\ifx \showCODEN    \undefined \def \showCODEN     #1{\unskip}     \fi
\ifx \showDOI      \undefined \def \showDOI       #1{#1}\fi
\ifx \showISBNx    \undefined \def \showISBNx     #1{\unskip}     \fi
\ifx \showISBNxiii \undefined \def \showISBNxiii  #1{\unskip}     \fi
\ifx \showISSN     \undefined \def \showISSN      #1{\unskip}     \fi
\ifx \showLCCN     \undefined \def \showLCCN      #1{\unskip}     \fi
\ifx \shownote     \undefined \def \shownote      #1{#1}          \fi
\ifx \showarticletitle \undefined \def \showarticletitle #1{#1}   \fi
\ifx \showURL      \undefined \def \showURL       {\relax}        \fi
\providecommand\bibfield[2]{#2}
\providecommand\bibinfo[2]{#2}
\providecommand\natexlab[1]{#1}
\providecommand\showeprint[2][]{arXiv:#2}

\bibitem[Aggarwal et~al\mbox{.}(2020)]%
        {aggarwal2020text}
\bibfield{author}{\bibinfo{person}{Surbhi Aggarwal}, \bibinfo{person}{Venkatesh~Babu Radhakrishnan}, {and} \bibinfo{person}{Anirban Chakraborty}.} \bibinfo{year}{2020}\natexlab{}.
\newblock \showarticletitle{Text-based person search via attribute-aided matching}. In \bibinfo{booktitle}{\emph{Proceedings of the IEEE/CVF winter conference on applications of computer vision}}. \bibinfo{pages}{2617--2625}.
\newblock


\bibitem[Anderson et~al\mbox{.}(2018)]%
        {anderson2018bottom}
\bibfield{author}{\bibinfo{person}{Peter Anderson}, \bibinfo{person}{Xiaodong He}, \bibinfo{person}{Chris Buehler}, \bibinfo{person}{Damien Teney}, \bibinfo{person}{Mark Johnson}, \bibinfo{person}{Stephen Gould}, {and} \bibinfo{person}{Lei Zhang}.} \bibinfo{year}{2018}\natexlab{}.
\newblock \showarticletitle{Bottom-up and top-down attention for image captioning and visual question answering}. In \bibinfo{booktitle}{\emph{Proceedings of the IEEE conference on computer vision and pattern recognition}}. \bibinfo{pages}{6077--6086}.
\newblock


\bibitem[Cascante-Bonilla et~al\mbox{.}(2022)]%
        {cascante2022simvqa}
\bibfield{author}{\bibinfo{person}{Paola Cascante-Bonilla}, \bibinfo{person}{Hui Wu}, \bibinfo{person}{Letao Wang}, \bibinfo{person}{Rogerio~S Feris}, {and} \bibinfo{person}{Vicente Ordonez}.} \bibinfo{year}{2022}\natexlab{}.
\newblock \showarticletitle{Simvqa: Exploring simulated environments for visual question answering}. In \bibinfo{booktitle}{\emph{Proceedings of the IEEE/CVF Conference on Computer Vision and Pattern Recognition}}. \bibinfo{pages}{5056--5066}.
\newblock


\bibitem[Chen et~al\mbox{.}(2023)]%
        {chen2023towards}
\bibfield{author}{\bibinfo{person}{Cuiqun Chen}, \bibinfo{person}{Mang Ye}, {and} \bibinfo{person}{Ding Jiang}.} \bibinfo{year}{2023}\natexlab{}.
\newblock \showarticletitle{Towards Modality-Agnostic Person Re-Identification With Descriptive Query}. In \bibinfo{booktitle}{\emph{Proceedings of the IEEE/CVF Conference on Computer Vision and Pattern Recognition}}. \bibinfo{pages}{15128--15137}.
\newblock


\bibitem[Chen et~al\mbox{.}(2018)]%
        {chen2018improving}
\bibfield{author}{\bibinfo{person}{Dapeng Chen}, \bibinfo{person}{Hongsheng Li}, \bibinfo{person}{Xihui Liu}, \bibinfo{person}{Yantao Shen}, \bibinfo{person}{Jing Shao}, \bibinfo{person}{Zejian Yuan}, {and} \bibinfo{person}{Xiaogang Wang}.} \bibinfo{year}{2018}\natexlab{}.
\newblock \showarticletitle{Improving deep visual representation for person re-identification by global and local image-language association}. In \bibinfo{booktitle}{\emph{Proceedings of the European conference on computer vision (ECCV)}}. \bibinfo{pages}{54--70}.
\newblock


\bibitem[Chen et~al\mbox{.}(2020)]%
        {chen2020imram}
\bibfield{author}{\bibinfo{person}{Hui Chen}, \bibinfo{person}{Guiguang Ding}, \bibinfo{person}{Xudong Liu}, \bibinfo{person}{Zijia Lin}, \bibinfo{person}{Ji Liu}, {and} \bibinfo{person}{Jungong Han}.} \bibinfo{year}{2020}\natexlab{}.
\newblock \showarticletitle{Imram: Iterative matching with recurrent attention memory for cross-modal image-text retrieval}. In \bibinfo{booktitle}{\emph{Proceedings of the IEEE/CVF conference on computer vision and pattern recognition}}. \bibinfo{pages}{12655--12663}.
\newblock


\bibitem[Chen and Luo(2020)]%
        {chen2020expressing}
\bibfield{author}{\bibinfo{person}{Tianlang Chen} {and} \bibinfo{person}{Jiebo Luo}.} \bibinfo{year}{2020}\natexlab{}.
\newblock \showarticletitle{Expressing objects just like words: Recurrent visual embedding for image-text matching}. In \bibinfo{booktitle}{\emph{Proceedings of the AAAI conference on artificial intelligence}}, Vol.~\bibinfo{volume}{34}. \bibinfo{pages}{10583--10590}.
\newblock


\bibitem[Chun et~al\mbox{.}(2021)]%
        {chun2021probabilistic}
\bibfield{author}{\bibinfo{person}{Sanghyuk Chun}, \bibinfo{person}{Seong~Joon Oh}, \bibinfo{person}{Rafael~Sampaio De~Rezende}, \bibinfo{person}{Yannis Kalantidis}, {and} \bibinfo{person}{Diane Larlus}.} \bibinfo{year}{2021}\natexlab{}.
\newblock \showarticletitle{Probabilistic embeddings for cross-modal retrieval}. In \bibinfo{booktitle}{\emph{Proceedings of the IEEE/CVF Conference on Computer Vision and Pattern Recognition}}. \bibinfo{pages}{8415--8424}.
\newblock


\bibitem[Deng et~al\mbox{.}(2009)]%
        {deng2009imagenet}
\bibfield{author}{\bibinfo{person}{Jia Deng}, \bibinfo{person}{Wei Dong}, \bibinfo{person}{Richard Socher}, \bibinfo{person}{Li-Jia Li}, \bibinfo{person}{Kai Li}, {and} \bibinfo{person}{Li Fei-Fei}.} \bibinfo{year}{2009}\natexlab{}.
\newblock \showarticletitle{Imagenet: A large-scale hierarchical image database}. In \bibinfo{booktitle}{\emph{2009 IEEE conference on computer vision and pattern recognition}}. Ieee, \bibinfo{pages}{248--255}.
\newblock


\bibitem[Devlin et~al\mbox{.}(2018)]%
        {devlin2018bert}
\bibfield{author}{\bibinfo{person}{Jacob Devlin}, \bibinfo{person}{Ming-Wei Chang}, \bibinfo{person}{Kenton Lee}, {and} \bibinfo{person}{Kristina Toutanova}.} \bibinfo{year}{2018}\natexlab{}.
\newblock \showarticletitle{Bert: Pre-training of deep bidirectional transformers for language understanding}.
\newblock \bibinfo{journal}{\emph{arXiv preprint arXiv:1810.04805}} (\bibinfo{year}{2018}).
\newblock


\bibitem[Ding and Mang(2023)]%
        {jiang_2023}
\bibfield{author}{\bibinfo{person}{Jiang Ding} {and} \bibinfo{person}{Ye Mang}.} \bibinfo{year}{2023}\natexlab{}.
\newblock \showarticletitle{Transformer Network for Cross-modal Text-to-Image Person Re-identification}.
\newblock \bibinfo{journal}{\emph{JOURNAL OF IMAGE AND GRAPHICS}} (\bibinfo{year}{2023}).
\newblock
\urldef\tempurl%
\url{https://doi.org/10.11834/jig.220620}
\showDOI{\tempurl}


\bibitem[Ding et~al\mbox{.}(2022)]%
        {ding2022mukea}
\bibfield{author}{\bibinfo{person}{Yang Ding}, \bibinfo{person}{Jing Yu}, \bibinfo{person}{Bang Liu}, \bibinfo{person}{Yue Hu}, \bibinfo{person}{Mingxin Cui}, {and} \bibinfo{person}{Qi Wu}.} \bibinfo{year}{2022}\natexlab{}.
\newblock \showarticletitle{MuKEA: Multimodal Knowledge Extraction and Accumulation for Knowledge-based Visual Question Answering}. In \bibinfo{booktitle}{\emph{Proceedings of the IEEE/CVF Conference on Computer Vision and Pattern Recognition}}. \bibinfo{pages}{5089--5098}.
\newblock


\bibitem[Ding et~al\mbox{.}(2021)]%
        {ding2021semantically}
\bibfield{author}{\bibinfo{person}{Zefeng Ding}, \bibinfo{person}{Changxing Ding}, \bibinfo{person}{Zhiyin Shao}, {and} \bibinfo{person}{Dacheng Tao}.} \bibinfo{year}{2021}\natexlab{}.
\newblock \showarticletitle{Semantically self-aligned network for text-to-image part-aware person re-identification}.
\newblock \bibinfo{journal}{\emph{arXiv preprint arXiv:2107.12666}} (\bibinfo{year}{2021}).
\newblock


\bibitem[Faghri et~al\mbox{.}(2017)]%
        {faghri2017vse++}
\bibfield{author}{\bibinfo{person}{Fartash Faghri}, \bibinfo{person}{David~J Fleet}, \bibinfo{person}{Jamie~Ryan Kiros}, {and} \bibinfo{person}{Sanja Fidler}.} \bibinfo{year}{2017}\natexlab{}.
\newblock \showarticletitle{Vse++: Improving visual-semantic embeddings with hard negatives}.
\newblock \bibinfo{journal}{\emph{arXiv preprint arXiv:1707.05612}} (\bibinfo{year}{2017}).
\newblock


\bibitem[Fang et~al\mbox{.}(2022)]%
        {fang2022injecting}
\bibfield{author}{\bibinfo{person}{Zhiyuan Fang}, \bibinfo{person}{Jianfeng Wang}, \bibinfo{person}{Xiaowei Hu}, \bibinfo{person}{Lin Liang}, \bibinfo{person}{Zhe Gan}, \bibinfo{person}{Lijuan Wang}, \bibinfo{person}{Yezhou Yang}, {and} \bibinfo{person}{Zicheng Liu}.} \bibinfo{year}{2022}\natexlab{}.
\newblock \showarticletitle{Injecting semantic concepts into end-to-end image captioning}. In \bibinfo{booktitle}{\emph{Proceedings of the IEEE/CVF Conference on Computer Vision and Pattern Recognition}}. \bibinfo{pages}{18009--18019}.
\newblock


\bibitem[Fei et~al\mbox{.}(2023)]%
        {fei-etal-2023-scene}
\bibfield{author}{\bibinfo{person}{Hao Fei}, \bibinfo{person}{Qian Liu}, \bibinfo{person}{Meishan Zhang}, \bibinfo{person}{Min Zhang}, {and} \bibinfo{person}{Tat-Seng Chua}.} \bibinfo{year}{2023}\natexlab{}.
\newblock \showarticletitle{Scene Graph as Pivoting: Inference-time Image-free Unsupervised Multimodal Machine Translation with Visual Scene Hallucination}. In \bibinfo{booktitle}{\emph{Proceedings of the 61st Annual Meeting of the Association for Computational Linguistics (Volume 1: Long Papers)}}. \bibinfo{pages}{5980--5994}.
\newblock


\bibitem[Fei et~al\mbox{.}(2022)]%
        {FeiMatchStruICML22}
\bibfield{author}{\bibinfo{person}{Hao Fei}, \bibinfo{person}{Shengqiong Wu}, \bibinfo{person}{Yafeng Ren}, {and} \bibinfo{person}{Meishan Zhang}.} \bibinfo{year}{2022}\natexlab{}.
\newblock \showarticletitle{Matching Structure for Dual Learning}. In \bibinfo{booktitle}{\emph{Proceedings of the International Conference on Machine Learning, {ICML}}}. \bibinfo{pages}{6373--6391}.
\newblock


\bibitem[Frome et~al\mbox{.}(2013)]%
        {frome2013devise}
\bibfield{author}{\bibinfo{person}{Andrea Frome}, \bibinfo{person}{Greg~S Corrado}, \bibinfo{person}{Jon Shlens}, \bibinfo{person}{Samy Bengio}, \bibinfo{person}{Jeff Dean}, \bibinfo{person}{Marc'Aurelio Ranzato}, {and} \bibinfo{person}{Tomas Mikolov}.} \bibinfo{year}{2013}\natexlab{}.
\newblock \showarticletitle{Devise: A deep visual-semantic embedding model}.
\newblock \bibinfo{journal}{\emph{Advances in neural information processing systems}}  \bibinfo{volume}{26} (\bibinfo{year}{2013}).
\newblock


\bibitem[Gao et~al\mbox{.}(2021)]%
        {gao2021contextual}
\bibfield{author}{\bibinfo{person}{Chenyang Gao}, \bibinfo{person}{Guanyu Cai}, \bibinfo{person}{Xinyang Jiang}, \bibinfo{person}{Feng Zheng}, \bibinfo{person}{Jun Zhang}, \bibinfo{person}{Yifei Gong}, \bibinfo{person}{Pai Peng}, \bibinfo{person}{Xiaowei Guo}, {and} \bibinfo{person}{Xing Sun}.} \bibinfo{year}{2021}\natexlab{}.
\newblock \showarticletitle{Contextual non-local alignment over full-scale representation for text-based person search}.
\newblock \bibinfo{journal}{\emph{arXiv preprint arXiv:2101.03036}} (\bibinfo{year}{2021}).
\newblock


\bibitem[Ge et~al\mbox{.}(2021)]%
        {ge2021structured}
\bibfield{author}{\bibinfo{person}{Xuri Ge}, \bibinfo{person}{Fuhai Chen}, \bibinfo{person}{Joemon~M Jose}, \bibinfo{person}{Zhilong Ji}, \bibinfo{person}{Zhongqin Wu}, {and} \bibinfo{person}{Xiao Liu}.} \bibinfo{year}{2021}\natexlab{}.
\newblock \showarticletitle{Structured multi-modal feature embedding and alignment for image-sentence retrieval}. In \bibinfo{booktitle}{\emph{Proceedings of the 29th ACM International Conference on Multimedia}}. \bibinfo{pages}{5185--5193}.
\newblock


\bibitem[Han et~al\mbox{.}(2021)]%
        {han2021text}
\bibfield{author}{\bibinfo{person}{Xiao Han}, \bibinfo{person}{Sen He}, \bibinfo{person}{Li Zhang}, {and} \bibinfo{person}{Tao Xiang}.} \bibinfo{year}{2021}\natexlab{}.
\newblock \showarticletitle{Text-based person search with limited data}.
\newblock \bibinfo{journal}{\emph{arXiv preprint arXiv:2110.10807}} (\bibinfo{year}{2021}).
\newblock


\bibitem[He et~al\mbox{.}(2016)]%
        {he2016deep}
\bibfield{author}{\bibinfo{person}{Kaiming He}, \bibinfo{person}{Xiangyu Zhang}, \bibinfo{person}{Shaoqing Ren}, {and} \bibinfo{person}{Jian Sun}.} \bibinfo{year}{2016}\natexlab{}.
\newblock \showarticletitle{Deep residual learning for image recognition}. In \bibinfo{booktitle}{\emph{Proceedings of the IEEE conference on computer vision and pattern recognition}}. \bibinfo{pages}{770--778}.
\newblock


\bibitem[Hu et~al\mbox{.}(2022)]%
        {Hu_2022_CVPR}
\bibfield{author}{\bibinfo{person}{Xiaowei Hu}, \bibinfo{person}{Zhe Gan}, \bibinfo{person}{Jianfeng Wang}, \bibinfo{person}{Zhengyuan Yang}, \bibinfo{person}{Zicheng Liu}, \bibinfo{person}{Yumao Lu}, {and} \bibinfo{person}{Lijuan Wang}.} \bibinfo{year}{2022}\natexlab{}.
\newblock \showarticletitle{Scaling Up Vision-Language Pre-Training for Image Captioning}. In \bibinfo{booktitle}{\emph{Proceedings of the IEEE/CVF Conference on Computer Vision and Pattern Recognition (CVPR)}}. \bibinfo{pages}{17980--17989}.
\newblock


\bibitem[Huang et~al\mbox{.}(2017)]%
        {huang2017instance}
\bibfield{author}{\bibinfo{person}{Yan Huang}, \bibinfo{person}{Wei Wang}, {and} \bibinfo{person}{Liang Wang}.} \bibinfo{year}{2017}\natexlab{}.
\newblock \showarticletitle{Instance-aware image and sentence matching with selective multimodal lstm}. In \bibinfo{booktitle}{\emph{Proceedings of the IEEE Conference on Computer Vision and Pattern Recognition}}. \bibinfo{pages}{2310--2318}.
\newblock


\bibitem[Ji et~al\mbox{.}(2022)]%
        {ji2022konowing}
\bibfield{author}{\bibinfo{person}{Jiayi Ji}, \bibinfo{person}{Yiwei Ma}, \bibinfo{person}{Xiaoshuai Sun}, \bibinfo{person}{Yiyi Zhou}, \bibinfo{person}{Yongjian Wu}, {and} \bibinfo{person}{Rongrong Ji}.} \bibinfo{year}{2022}\natexlab{}.
\newblock \showarticletitle{Knowing What to Learn: A Metric-Oriented Focal Mechanism for Image Captioning}.
\newblock \bibinfo{journal}{\emph{IEEE Transactions on Image Processing}}  \bibinfo{volume}{31} (\bibinfo{year}{2022}), \bibinfo{pages}{4321--4335}.
\newblock
\urldef\tempurl%
\url{https://doi.org/10.1109/TIP.2022.3183434}
\showDOI{\tempurl}


\bibitem[Ji et~al\mbox{.}(2021)]%
        {ji2021step}
\bibfield{author}{\bibinfo{person}{Zhong Ji}, \bibinfo{person}{Kexin Chen}, {and} \bibinfo{person}{Haoran Wang}.} \bibinfo{year}{2021}\natexlab{}.
\newblock \showarticletitle{Step-wise hierarchical alignment network for image-text matching}.
\newblock \bibinfo{journal}{\emph{IJCAI}} (\bibinfo{year}{2021}).
\newblock


\bibitem[Jiang and Ye(2023)]%
        {jiang2023cross}
\bibfield{author}{\bibinfo{person}{Ding Jiang} {and} \bibinfo{person}{Mang Ye}.} \bibinfo{year}{2023}\natexlab{}.
\newblock \showarticletitle{Cross-Modal Implicit Relation Reasoning and Aligning for Text-to-Image Person Retrieval}. In \bibinfo{booktitle}{\emph{Proceedings of the IEEE/CVF Conference on Computer Vision and Pattern Recognition}}. \bibinfo{pages}{2787--2797}.
\newblock


\bibitem[Jiang et~al\mbox{.}(2020)]%
        {jiang2020defense}
\bibfield{author}{\bibinfo{person}{Huaizu Jiang}, \bibinfo{person}{Ishan Misra}, \bibinfo{person}{Marcus Rohrbach}, \bibinfo{person}{Erik Learned-Miller}, {and} \bibinfo{person}{Xinlei Chen}.} \bibinfo{year}{2020}\natexlab{}.
\newblock \showarticletitle{In defense of grid features for visual question answering}. In \bibinfo{booktitle}{\emph{Proceedings of the IEEE/CVF Conference on Computer Vision and Pattern Recognition}}. \bibinfo{pages}{10267--10276}.
\newblock


\bibitem[Jing et~al\mbox{.}(2022)]%
        {Jing_2022_CVPR}
\bibfield{author}{\bibinfo{person}{Chenchen Jing}, \bibinfo{person}{Yunde Jia}, \bibinfo{person}{Yuwei Wu}, \bibinfo{person}{Xinyu Liu}, {and} \bibinfo{person}{Qi Wu}.} \bibinfo{year}{2022}\natexlab{}.
\newblock \showarticletitle{Maintaining Reasoning Consistency in Compositional Visual Question Answering}. In \bibinfo{booktitle}{\emph{Proceedings of the IEEE/CVF Conference on Computer Vision and Pattern Recognition (CVPR)}}. \bibinfo{pages}{5099--5108}.
\newblock


\bibitem[Jing et~al\mbox{.}(2020a)]%
        {jing2020pose}
\bibfield{author}{\bibinfo{person}{Ya Jing}, \bibinfo{person}{Chenyang Si}, \bibinfo{person}{Junbo Wang}, \bibinfo{person}{Wei Wang}, \bibinfo{person}{Liang Wang}, {and} \bibinfo{person}{Tieniu Tan}.} \bibinfo{year}{2020}\natexlab{a}.
\newblock \showarticletitle{Pose-guided multi-granularity attention network for text-based person search}. In \bibinfo{booktitle}{\emph{Proceedings of the AAAI Conference on Artificial Intelligence}}, Vol.~\bibinfo{volume}{34}. \bibinfo{pages}{11189--11196}.
\newblock


\bibitem[Jing et~al\mbox{.}(2020b)]%
        {jing2020cross}
\bibfield{author}{\bibinfo{person}{Ya Jing}, \bibinfo{person}{Wei Wang}, \bibinfo{person}{Liang Wang}, {and} \bibinfo{person}{Tieniu Tan}.} \bibinfo{year}{2020}\natexlab{b}.
\newblock \showarticletitle{Cross-modal cross-domain moment alignment network for person search}. In \bibinfo{booktitle}{\emph{Proceedings of the IEEE/CVF Conference on Computer Vision and Pattern Recognition}}. \bibinfo{pages}{10678--10686}.
\newblock


\bibitem[Kingma and Ba(2014)]%
        {kingma2014adam}
\bibfield{author}{\bibinfo{person}{Diederik~P Kingma} {and} \bibinfo{person}{Jimmy Ba}.} \bibinfo{year}{2014}\natexlab{}.
\newblock \showarticletitle{Adam: A method for stochastic optimization}.
\newblock \bibinfo{journal}{\emph{arXiv preprint arXiv:1412.6980}} (\bibinfo{year}{2014}).
\newblock


\bibitem[Lee et~al\mbox{.}(2018)]%
        {lee2018stacked}
\bibfield{author}{\bibinfo{person}{Kuang-Huei Lee}, \bibinfo{person}{Xi Chen}, \bibinfo{person}{Gang Hua}, \bibinfo{person}{Houdong Hu}, {and} \bibinfo{person}{Xiaodong He}.} \bibinfo{year}{2018}\natexlab{}.
\newblock \showarticletitle{Stacked cross attention for image-text matching}. In \bibinfo{booktitle}{\emph{Proceedings of the European conference on computer vision (ECCV)}}. \bibinfo{pages}{201--216}.
\newblock


\bibitem[Li et~al\mbox{.}(2017a)]%
        {li2017identity}
\bibfield{author}{\bibinfo{person}{Shuang Li}, \bibinfo{person}{Tong Xiao}, \bibinfo{person}{Hongsheng Li}, \bibinfo{person}{Wei Yang}, {and} \bibinfo{person}{Xiaogang Wang}.} \bibinfo{year}{2017}\natexlab{a}.
\newblock \showarticletitle{Identity-aware textual-visual matching with latent co-attention}. In \bibinfo{booktitle}{\emph{Proceedings of the IEEE International Conference on Computer Vision}}. \bibinfo{pages}{1890--1899}.
\newblock


\bibitem[Li et~al\mbox{.}(2017b)]%
        {li2017person}
\bibfield{author}{\bibinfo{person}{Shuang Li}, \bibinfo{person}{Tong Xiao}, \bibinfo{person}{Hongsheng Li}, \bibinfo{person}{Bolei Zhou}, \bibinfo{person}{Dayu Yue}, {and} \bibinfo{person}{Xiaogang Wang}.} \bibinfo{year}{2017}\natexlab{b}.
\newblock \showarticletitle{Person search with natural language description}. In \bibinfo{booktitle}{\emph{Proceedings of the IEEE Conference on Computer Vision and Pattern Recognition}}. \bibinfo{pages}{1970--1979}.
\newblock


\bibitem[Lin et~al\mbox{.}(2020)]%
        {lin2020many}
\bibfield{author}{\bibinfo{person}{Jialiang Lin}, \bibinfo{person}{Yao Yu}, \bibinfo{person}{Yu Zhou}, \bibinfo{person}{Zhiyang Zhou}, {and} \bibinfo{person}{Xiaodong Shi}.} \bibinfo{year}{2020}\natexlab{}.
\newblock \showarticletitle{How many preprints have actually been printed and why: a case study of computer science preprints on arXiv}.
\newblock \bibinfo{journal}{\emph{Scientometrics}} \bibinfo{volume}{124}, \bibinfo{number}{1} (\bibinfo{year}{2020}), \bibinfo{pages}{555--574}.
\newblock


\bibitem[Lin et~al\mbox{.}(2014)]%
        {lin2014microsoft}
\bibfield{author}{\bibinfo{person}{Tsung-Yi Lin}, \bibinfo{person}{Michael Maire}, \bibinfo{person}{Serge Belongie}, \bibinfo{person}{James Hays}, \bibinfo{person}{Pietro Perona}, \bibinfo{person}{Deva Ramanan}, \bibinfo{person}{Piotr Doll{\'a}r}, {and} \bibinfo{person}{C~Lawrence Zitnick}.} \bibinfo{year}{2014}\natexlab{}.
\newblock \showarticletitle{Microsoft coco: Common objects in context}. In \bibinfo{booktitle}{\emph{European conference on computer vision}}. Springer, \bibinfo{pages}{740--755}.
\newblock


\bibitem[Liu et~al\mbox{.}(2019a)]%
        {liu2019focus}
\bibfield{author}{\bibinfo{person}{Chunxiao Liu}, \bibinfo{person}{Zhendong Mao}, \bibinfo{person}{An-An Liu}, \bibinfo{person}{Tianzhu Zhang}, \bibinfo{person}{Bin Wang}, {and} \bibinfo{person}{Yongdong Zhang}.} \bibinfo{year}{2019}\natexlab{a}.
\newblock \showarticletitle{Focus your attention: A bidirectional focal attention network for image-text matching}. In \bibinfo{booktitle}{\emph{Proceedings of the 27th ACM International Conference on Multimedia}}. \bibinfo{pages}{3--11}.
\newblock


\bibitem[Liu et~al\mbox{.}(2019b)]%
        {liu2019deep}
\bibfield{author}{\bibinfo{person}{Jiawei Liu}, \bibinfo{person}{Zheng-Jun Zha}, \bibinfo{person}{Richang Hong}, \bibinfo{person}{Meng Wang}, {and} \bibinfo{person}{Yongdong Zhang}.} \bibinfo{year}{2019}\natexlab{b}.
\newblock \showarticletitle{Deep adversarial graph attention convolution network for text-based person search}. In \bibinfo{booktitle}{\emph{Proceedings of the 27th ACM International Conference on Multimedia}}. \bibinfo{pages}{665--673}.
\newblock


\bibitem[Ma et~al\mbox{.}(2022)]%
        {ma2022knowing}
\bibfield{author}{\bibinfo{person}{Yiwei Ma}, \bibinfo{person}{Jiayi Ji}, \bibinfo{person}{Xiaoshuai Sun}, \bibinfo{person}{Yiyi Zhou}, \bibinfo{person}{Yongjian Wu}, \bibinfo{person}{Feiyue Huang}, {and} \bibinfo{person}{Rongrong Ji}.} \bibinfo{year}{2022}\natexlab{}.
\newblock \showarticletitle{Knowing what it is: Semantic-enhanced Dual Attention Transformer}.
\newblock \bibinfo{journal}{\emph{IEEE Transactions on Multimedia}} (\bibinfo{year}{2022}), \bibinfo{pages}{1--1}.
\newblock
\urldef\tempurl%
\url{https://doi.org/10.1109/TMM.2022.3164787}
\showDOI{\tempurl}


\bibitem[Niu et~al\mbox{.}(2020)]%
        {niu2020improving}
\bibfield{author}{\bibinfo{person}{Kai Niu}, \bibinfo{person}{Yan Huang}, \bibinfo{person}{Wanli Ouyang}, {and} \bibinfo{person}{Liang Wang}.} \bibinfo{year}{2020}\natexlab{}.
\newblock \showarticletitle{Improving description-based person re-identification by multi-granularity image-text alignments}.
\newblock \bibinfo{journal}{\emph{IEEE Transactions on Image Processing}}  \bibinfo{volume}{29} (\bibinfo{year}{2020}), \bibinfo{pages}{5542--5556}.
\newblock


\bibitem[Reed et~al\mbox{.}(2016)]%
        {reed2016learning}
\bibfield{author}{\bibinfo{person}{Scott Reed}, \bibinfo{person}{Zeynep Akata}, \bibinfo{person}{Honglak Lee}, {and} \bibinfo{person}{Bernt Schiele}.} \bibinfo{year}{2016}\natexlab{}.
\newblock \showarticletitle{Learning deep representations of fine-grained visual descriptions}. In \bibinfo{booktitle}{\emph{Proceedings of the IEEE conference on computer vision and pattern recognition}}. \bibinfo{pages}{49--58}.
\newblock


\bibitem[Shao et~al\mbox{.}(2022)]%
        {shao2022learning}
\bibfield{author}{\bibinfo{person}{Zhiyin Shao}, \bibinfo{person}{Xinyu Zhang}, \bibinfo{person}{Meng Fang}, \bibinfo{person}{Zhifeng Lin}, \bibinfo{person}{Jian Wang}, {and} \bibinfo{person}{Changxing Ding}.} \bibinfo{year}{2022}\natexlab{}.
\newblock \showarticletitle{Learning Granularity-Unified Representations for Text-to-Image Person Re-identification}.
\newblock \bibinfo{journal}{\emph{arXiv preprint arXiv:2207.07802}} (\bibinfo{year}{2022}).
\newblock


\bibitem[Song and Soleymani(2019)]%
        {song2019polysemous}
\bibfield{author}{\bibinfo{person}{Yale Song} {and} \bibinfo{person}{Mohammad Soleymani}.} \bibinfo{year}{2019}\natexlab{}.
\newblock \showarticletitle{Polysemous visual-semantic embedding for cross-modal retrieval}. In \bibinfo{booktitle}{\emph{Proceedings of the IEEE/CVF Conference on Computer Vision and Pattern Recognition}}. \bibinfo{pages}{1979--1988}.
\newblock


\bibitem[Van~der Maaten and Hinton(2008)]%
        {van2008visualizing}
\bibfield{author}{\bibinfo{person}{Laurens Van~der Maaten} {and} \bibinfo{person}{Geoffrey Hinton}.} \bibinfo{year}{2008}\natexlab{}.
\newblock \showarticletitle{Visualizing data using t-SNE.}
\newblock \bibinfo{journal}{\emph{Journal of machine learning research}} \bibinfo{volume}{9}, \bibinfo{number}{11} (\bibinfo{year}{2008}).
\newblock


\bibitem[Wang et~al\mbox{.}(2021a)]%
        {wang2021text}
\bibfield{author}{\bibinfo{person}{Chengji Wang}, \bibinfo{person}{Zhiming Luo}, \bibinfo{person}{Yaojin Lin}, {and} \bibinfo{person}{Shaozi Li}.} \bibinfo{year}{2021}\natexlab{a}.
\newblock \showarticletitle{Text-based Person Search via Multi-Granularity Embedding Learning.}. In \bibinfo{booktitle}{\emph{IJCAI}}. \bibinfo{pages}{1068--1074}.
\newblock


\bibitem[Wang et~al\mbox{.}(2023b)]%
        {wang2023exploiting}
\bibfield{author}{\bibinfo{person}{Guanshuo Wang}, \bibinfo{person}{Fufu Yu}, \bibinfo{person}{Junjie Li}, \bibinfo{person}{Qiong Jia}, {and} \bibinfo{person}{Shouhong Ding}.} \bibinfo{year}{2023}\natexlab{b}.
\newblock \showarticletitle{Exploiting the Textual Potential from Vision-Language Pre-training for Text-based Person Search}.
\newblock \bibinfo{journal}{\emph{arXiv preprint arXiv:2303.04497}} (\bibinfo{year}{2023}).
\newblock


\bibitem[Wang et~al\mbox{.}(2023a)]%
        {wang2023towards}
\bibfield{author}{\bibinfo{person}{Haowei Wang}, \bibinfo{person}{Jiayi Ji}, \bibinfo{person}{Yiyi Zhou}, \bibinfo{person}{Yongjian Wu}, {and} \bibinfo{person}{Xiaoshuai Sun}.} \bibinfo{year}{2023}\natexlab{a}.
\newblock \showarticletitle{Towards real-time panoptic narrative grounding by an end-to-end grounding network}.
\newblock \bibinfo{journal}{\emph{arXiv preprint arXiv:2301.03160}} (\bibinfo{year}{2023}).
\newblock


\bibitem[Wang et~al\mbox{.}(2020c)]%
        {wang2020consensus}
\bibfield{author}{\bibinfo{person}{Haoran Wang}, \bibinfo{person}{Ying Zhang}, \bibinfo{person}{Zhong Ji}, \bibinfo{person}{Yanwei Pang}, {and} \bibinfo{person}{Lin Ma}.} \bibinfo{year}{2020}\natexlab{c}.
\newblock \showarticletitle{Consensus-aware visual-semantic embedding for image-text matching}. In \bibinfo{booktitle}{\emph{European Conference on Computer Vision}}. Springer, \bibinfo{pages}{18--34}.
\newblock


\bibitem[Wang et~al\mbox{.}(2020b)]%
        {wang2020cross}
\bibfield{author}{\bibinfo{person}{Sijin Wang}, \bibinfo{person}{Ruiping Wang}, \bibinfo{person}{Ziwei Yao}, \bibinfo{person}{Shiguang Shan}, {and} \bibinfo{person}{Xilin Chen}.} \bibinfo{year}{2020}\natexlab{b}.
\newblock \showarticletitle{Cross-modal scene graph matching for relationship-aware image-text retrieval}. In \bibinfo{booktitle}{\emph{Proceedings of the IEEE/CVF winter conference on applications of computer vision}}. \bibinfo{pages}{1508--1517}.
\newblock


\bibitem[Wang et~al\mbox{.}(2020a)]%
        {wang2020vitaa}
\bibfield{author}{\bibinfo{person}{Zhe Wang}, \bibinfo{person}{Zhiyuan Fang}, \bibinfo{person}{Jun Wang}, {and} \bibinfo{person}{Yezhou Yang}.} \bibinfo{year}{2020}\natexlab{a}.
\newblock \showarticletitle{Vitaa: Visual-textual attributes alignment in person search by natural language}. In \bibinfo{booktitle}{\emph{European Conference on Computer Vision}}. Springer, \bibinfo{pages}{402--420}.
\newblock


\bibitem[Wang et~al\mbox{.}(2022a)]%
        {wang2022aspd}
\bibfield{author}{\bibinfo{person}{Zijie Wang}, \bibinfo{person}{Jingyi Xue}, \bibinfo{person}{Xili Wan}, \bibinfo{person}{Aichun Zhu}, \bibinfo{person}{Yifeng Li}, \bibinfo{person}{Xiaomei Zhu}, {and} \bibinfo{person}{Fangqiang Hu}.} \bibinfo{year}{2022}\natexlab{a}.
\newblock \showarticletitle{ASPD-Net: Self-aligned part mask for improving text-based person re-identification with adversarial representation learning}.
\newblock \bibinfo{journal}{\emph{Engineering Applications of Artificial Intelligence}}  \bibinfo{volume}{116} (\bibinfo{year}{2022}), \bibinfo{pages}{105419}.
\newblock


\bibitem[Wang et~al\mbox{.}(2021b)]%
        {wang2021amen}
\bibfield{author}{\bibinfo{person}{Zijie Wang}, \bibinfo{person}{Jingyi Xue}, \bibinfo{person}{Aichun Zhu}, \bibinfo{person}{Yifeng Li}, \bibinfo{person}{Mingyi Zhang}, {and} \bibinfo{person}{Chongliang Zhong}.} \bibinfo{year}{2021}\natexlab{b}.
\newblock \showarticletitle{AMEN: Adversarial Multi-space Embedding Network for Text-Based Person Re-identification}. In \bibinfo{booktitle}{\emph{Chinese Conference on Pattern Recognition and Computer Vision (PRCV)}}. Springer, \bibinfo{pages}{462--473}.
\newblock


\bibitem[Wang et~al\mbox{.}(2022b)]%
        {wang2022sum}
\bibfield{author}{\bibinfo{person}{Zijie Wang}, \bibinfo{person}{Aichun Zhu}, \bibinfo{person}{Jingyi Xue}, \bibinfo{person}{Daihong Jiang}, \bibinfo{person}{Chao Liu}, \bibinfo{person}{Yifeng Li}, {and} \bibinfo{person}{Fangqiang Hu}.} \bibinfo{year}{2022}\natexlab{b}.
\newblock \showarticletitle{SUM: Serialized Updating and Matching for text-based person retrieval}.
\newblock \bibinfo{journal}{\emph{Knowledge-Based Systems}}  \bibinfo{volume}{248} (\bibinfo{year}{2022}), \bibinfo{pages}{108891}.
\newblock


\bibitem[Wang et~al\mbox{.}(2022c)]%
        {wang2022caibc}
\bibfield{author}{\bibinfo{person}{Zijie Wang}, \bibinfo{person}{Aichun Zhu}, \bibinfo{person}{Jingyi Xue}, \bibinfo{person}{Xili Wan}, \bibinfo{person}{Chao Liu}, \bibinfo{person}{Tian Wang}, {and} \bibinfo{person}{Yifeng Li}.} \bibinfo{year}{2022}\natexlab{c}.
\newblock \showarticletitle{CAIBC: Capturing All-round Information Beyond Color for Text-based Person Retrieval}.
\newblock \bibinfo{journal}{\emph{arXiv preprint arXiv:2209.05773}} (\bibinfo{year}{2022}).
\newblock


\bibitem[Wang et~al\mbox{.}(2022d)]%
        {wang2022look}
\bibfield{author}{\bibinfo{person}{Zijie Wang}, \bibinfo{person}{Aichun Zhu}, \bibinfo{person}{Jingyi Xue}, \bibinfo{person}{Xili Wan}, \bibinfo{person}{Chao Liu}, \bibinfo{person}{Tian Wang}, {and} \bibinfo{person}{Yifeng Li}.} \bibinfo{year}{2022}\natexlab{d}.
\newblock \showarticletitle{Look before you leap: Improving text-based person retrieval by learning a consistent cross-modal common manifold}. In \bibinfo{booktitle}{\emph{Proceedings of the 30th ACM International Conference on Multimedia}}. \bibinfo{pages}{1984--1992}.
\newblock


\bibitem[Wang et~al\mbox{.}(2020d)]%
        {wang2020img}
\bibfield{author}{\bibinfo{person}{Zijie Wang}, \bibinfo{person}{Aichun Zhu}, \bibinfo{person}{Zhe Zheng}, \bibinfo{person}{Jing Jin}, \bibinfo{person}{Zhouxin Xue}, {and} \bibinfo{person}{Gang Hua}.} \bibinfo{year}{2020}\natexlab{d}.
\newblock \showarticletitle{IMG-Net: inner-cross-modal attentional multigranular network for description-based person re-identification}.
\newblock \bibinfo{journal}{\emph{Journal of Electronic Imaging}} \bibinfo{volume}{29}, \bibinfo{number}{4} (\bibinfo{year}{2020}), \bibinfo{pages}{043028}.
\newblock


\bibitem[Wu et~al\mbox{.}(2021)]%
        {wu2021lapscore}
\bibfield{author}{\bibinfo{person}{Yushuang Wu}, \bibinfo{person}{Zizheng Yan}, \bibinfo{person}{Xiaoguang Han}, \bibinfo{person}{Guanbin Li}, \bibinfo{person}{Changqing Zou}, {and} \bibinfo{person}{Shuguang Cui}.} \bibinfo{year}{2021}\natexlab{}.
\newblock \showarticletitle{LapsCore: Language-guided Person Search via Color Reasoning}. In \bibinfo{booktitle}{\emph{Proceedings of the IEEE/CVF International Conference on Computer Vision}}. \bibinfo{pages}{1624--1633}.
\newblock


\bibitem[Xu et~al\mbox{.}(2023)]%
        {xu2023mining}
\bibfield{author}{\bibinfo{person}{Wenhao Xu}, \bibinfo{person}{Zhiyin Shao}, {and} \bibinfo{person}{Changxing Ding}.} \bibinfo{year}{2023}\natexlab{}.
\newblock \showarticletitle{Mining False Positive Examples for Text-Based Person Re-identification}.
\newblock \bibinfo{journal}{\emph{arXiv preprint arXiv:2303.08466}} (\bibinfo{year}{2023}).
\newblock


\bibitem[Zhang et~al\mbox{.}(2022)]%
        {zhang2022negative}
\bibfield{author}{\bibinfo{person}{Kun Zhang}, \bibinfo{person}{Zhendong Mao}, \bibinfo{person}{Quan Wang}, {and} \bibinfo{person}{Yongdong Zhang}.} \bibinfo{year}{2022}\natexlab{}.
\newblock \showarticletitle{Negative-Aware Attention Framework for Image-Text Matching}. In \bibinfo{booktitle}{\emph{Proceedings of the IEEE/CVF Conference on Computer Vision and Pattern Recognition}}. \bibinfo{pages}{15661--15670}.
\newblock


\bibitem[Zhang and Lu(2018)]%
        {zhang2018deep}
\bibfield{author}{\bibinfo{person}{Ying Zhang} {and} \bibinfo{person}{Huchuan Lu}.} \bibinfo{year}{2018}\natexlab{}.
\newblock \showarticletitle{Deep cross-modal projection learning for image-text matching}. In \bibinfo{booktitle}{\emph{Proceedings of the European conference on computer vision (ECCV)}}. \bibinfo{pages}{686--701}.
\newblock


\bibitem[Zheng et~al\mbox{.}(2020a)]%
        {zheng2020hierarchical}
\bibfield{author}{\bibinfo{person}{Kecheng Zheng}, \bibinfo{person}{Wu Liu}, \bibinfo{person}{Jiawei Liu}, \bibinfo{person}{Zheng-Jun Zha}, {and} \bibinfo{person}{Tao Mei}.} \bibinfo{year}{2020}\natexlab{a}.
\newblock \showarticletitle{Hierarchical gumbel attention network for text-based person search}. In \bibinfo{booktitle}{\emph{Proceedings of the 28th ACM International Conference on Multimedia}}. \bibinfo{pages}{3441--3449}.
\newblock


\bibitem[Zheng et~al\mbox{.}(2020b)]%
        {zheng2020dual}
\bibfield{author}{\bibinfo{person}{Zhedong Zheng}, \bibinfo{person}{Liang Zheng}, \bibinfo{person}{Michael Garrett}, \bibinfo{person}{Yi Yang}, \bibinfo{person}{Mingliang Xu}, {and} \bibinfo{person}{Yi-Dong Shen}.} \bibinfo{year}{2020}\natexlab{b}.
\newblock \showarticletitle{Dual-path convolutional image-text embeddings with instance loss}.
\newblock \bibinfo{journal}{\emph{ACM Transactions on Multimedia Computing, Communications, and Applications (TOMM)}} \bibinfo{volume}{16}, \bibinfo{number}{2} (\bibinfo{year}{2020}), \bibinfo{pages}{1--23}.
\newblock


\bibitem[Zhu et~al\mbox{.}(2021)]%
        {zhu2021dssl}
\bibfield{author}{\bibinfo{person}{Aichun Zhu}, \bibinfo{person}{Zijie Wang}, \bibinfo{person}{Yifeng Li}, \bibinfo{person}{Xili Wan}, \bibinfo{person}{Jing Jin}, \bibinfo{person}{Tian Wang}, \bibinfo{person}{Fangqiang Hu}, {and} \bibinfo{person}{Gang Hua}.} \bibinfo{year}{2021}\natexlab{}.
\newblock \showarticletitle{DSSL: Deep Surroundings-person Separation Learning for Text-based Person Retrieval}. In \bibinfo{booktitle}{\emph{Proceedings of the 29th ACM International Conference on Multimedia}}. \bibinfo{pages}{209--217}.
\newblock


\end{thebibliography}

\clearpage
\balance


\end{document}